\documentclass{article}

\usepackage[final]{neurips_2025}

\usepackage[utf8]{inputenc} 
\usepackage[T1]{fontenc}    
\usepackage{hyperref}       
\usepackage{url}            
\usepackage{booktabs}       
\usepackage{amsfonts}       
\usepackage{nicefrac}       
\usepackage{microtype}      
\usepackage{xcolor}         
\usepackage{graphicx}
\usepackage{subcaption} 
\usepackage{wrapfig}
\usepackage{booktabs}  

\usepackage{tcolorbox}
\definecolor{darkmagenta}{rgb}{0.56, 0.0, 1.0}  

\hypersetup{colorlinks,linkcolor={blue},citecolor={darkmagenta},urlcolor={blue}} 

\usepackage{soul}      
\usepackage{xcolor}
\definecolor{pastelSmall}{RGB}{245, 230, 184}
\definecolor{pastelMedium}{RGB}{244, 196, 200}
\definecolor{pastelLarge}{RGB}{207, 201, 232}

\colorlet{pastelMediumLight}{pastelMedium!45!white}  

\newcommand{\hlSmall}[1]{\sethlcolor{pastelSmall}\hl{#1}}
\newcommand{\hlMedium}[1]{\sethlcolor{pastelMedium}\hl{#1}}

\usepackage[table]{xcolor}
\usepackage{booktabs}
\usepackage{siunitx}

\definecolor{WidthSmall}{HTML}{E3C67E}  
\definecolor{WidthMedium}{HTML}{D4627D} 
\definecolor{WidthLarge}{HTML}{3B2073}  

\title{Stable Gradients for Stable Learning at Scale in Deep Reinforcement Learning}

\author{
  Roger Creus Castanyer\textsuperscript{1,2}\thanks{Equal contribution.  Correspondence to: Roger Creus C <\texttt{roger.creus-castanyer@mila.quebec}>, \\
  Johan Obando-Ceron <  \texttt{jobando0730@gmail.com}>,
  Pablo Samuel Castro <\texttt{psc@google.com}>}
 \quad
  Johan Obando-Ceron\textsuperscript{1,2}\footnotemark[1] \quad
  Lu Li\textsuperscript{1,2} \quad
  Pierre-Luc Bacon\textsuperscript{1,2} \quad \\
  \textbf{Glen Berseth\textsuperscript{1,2} \quad
  Aaron Courville\textsuperscript{1,2} \quad
  Pablo Samuel Castro\textsuperscript{1,2,3}} \\\\
  \textsuperscript{1} Mila - Qu\'ebec AI Institute \quad
  \textsuperscript{2} Universit\'e de Montr\'eal  \quad
  \textsuperscript{3} Google DeepMind \quad 
}

\begin{document}

\maketitle

\begin{abstract}
Scaling deep reinforcement learning networks is challenging and often results in degraded performance, yet the root causes of this failure mode remain poorly understood. Several recent works have proposed mechanisms to address this, but they are often complex and fail to highlight the causes underlying this difficulty. In this work, we conduct a series of empirical analyses which suggest that the combination of non-stationarity with gradient pathologies, due to suboptimal architectural choices, underlie the challenges of scale. We propose a series of direct interventions that stabilize gradient flow, enabling robust performance across a range of network depths and widths. Our interventions are simple to implement and compatible with well-established algorithms, and result in an effective mechanism that enables strong performance even at large scales. We validate our findings on a variety of agents and suites of environments. \href{https://github.com/roger-creus/stable-deep-rl-at-scale}{\textbf{Source code here.}}
\end{abstract}
\vspace{-1.3em}

\begin{center}
\begin{minipage}{0.9\linewidth}
\itshape
``We must be able to look at the world and see it as a dynamic process, not a static picture.''

\hfill --- David Bohm
\end{minipage}
\end{center}

\section{Introduction}
\label{sec:introduction}

Recent advances in deep reinforcement learning (deep RL) have demonstrated the ability of deep neural networks to solve complex decision-making tasks from robotics to game play and resource optimization \citep{mnih2015humanlevel, vinyals2019grandmaster,Bellemare2020AutonomousNO,fawzi2022discovering}. Motivated by successes in supervised and generative learning, recent works have explored scaling architectures in deep RL, showing gains in representation quality and generalization across tasks \citep{farebrother2023protovalue, taiga2023investigating}. 
However, scaling neural networks in deep RL remains fundamentally challenging \citep{ceron2024mixtures,ceron2024value}. 
A central cause of this instability lies in the unique optimization challenges of RL. Unlike supervised learning, where data distributions are fixed, deep RL involves policy-dependent data that constantly change during training \citep{lyle2022understanding}. Each update of the policy $\pi_\theta$ alters future states and rewards, making the training objective inherently \textit{non-stationary.}  
Value-based methods exacerbate these issues via bootstrapping, recursively using predicted values as targets. 

Estimation errors compound over time \citep{fujimoto2018addressing}, especially under sparse or delayed rewards \citep{zheng2018learning}, leading to unstable updates, policy collapse, or value divergence \citep{van2016deep,lyle2023understanding,lyle2024normalization}. 
These dynamics are tightly coupled with architectural vulnerabilities. Deep networks face well known pathologies such as vanishing/exploding gradients \citep{pascanu2013difficulty}, ill-conditioned Jacobians \citep{pennington2017resurrecting}, and activation saturation \citep{glorot2010understanding}. In deep RL, these are magnified by the ``\textit{deadly triad}'' \citep{sutton2018,van2018deep}, off-policy corrections, and changing targets. As networks scale, the risk of signal distortion and misalignment increases, resulting in underutilized capacity and brittle learning \citep{obando2023small, ceron2024value}.

One overlooked source of these failures lies in how gradients propagate through the network. Specifically, the gradient decomposition, the layer-wise structure of backpropagation as a chain of Jacobians and weights determine how information flows during learning \citep{Lee2020A}. 
While gradient signal preservation has been studied in supervised learning \citep{schoenholz2017deep, jacot2018neural}, its role in deep RL, where both inputs and targets shift continually, remains poorly understood.

In this work, we investigate how gradient decomposition interacts with non-stationarity and network scaling in deep RL. We demonstrate that in non-stationary settings like RL -- where targets are bootstrapped, policies evolve continually, and data distributions shift -- gradient signals progressively degrade across depth. This motivates the need for methods that explicitly preserve the structure of gradient information across layers. We explore this through a series of controlled experiments and ablations across multiple algorithms and environments, demonstrating that actively encouraging gradient propagation significantly improves stability and performance, even with large networks. Our work offers a promising approach for scaling deep RL architectures, yielding substantial performance gains across a variety of agents and training regimes.
\label{sec:stabilizingGradients}
\section{Preliminaries}
\label{sec:preliminaries}

\paragraph{Deep Reinforcement Learning} A deep reinforcement learning agent interacts with an environment through sequences of actions ($a\in\mathcal{A}$), which produce corresponding sequences of observations ($s\in\mathcal{S}$) and rewards ($r\in\mathbb{R}$), resulting in trajectories of the form $\tau := \{ s_0, a_0, r_0, s_1, a_1, r_2, \ldots \}$. The agent's behavior is often represented by a neural network with parameters $\theta$, composed of convolutional layers $\{\phi_1, \phi_2, \dots, \phi_{L_c}\}$ and dense (fully connected) layers $\{\psi_1, \psi_2, \dots, \psi_{L_d}\}$, where $\psi_{L_d}$ has an output dimensionality of $|\mathcal{A}|$. 
At every timestep $t$, an observation $s_t \in \mathcal{S}$ is fed through the network to obtain an estimate of the long-term value of each action: $Q_{\theta}(s_t, \cdot) = \psi_{L_d}(\psi_{L_d-1}(\ldots (\phi_{L_c}(\ldots (\phi_1(s_t))\ldots ))\ldots ))$.
The agent's policy $\pi_{\theta}(\cdot \mid s_t)$ specifies the probability of selecting each action, for instance by taking the softmax over the estimated values as in \autoref{equa:1}.
The training objective is typically defined as the maximization of expected cumulative reward as in \autoref{equa:2},

\begin{minipage}[t]{0.5\textwidth}
\begin{equation}
\label{equa:1}
\pi_{\theta}(a_t \mid s_t) = \frac{e^{Q_{\theta}(s_t, a_t)}}{\sum_{a\in\mathcal{A}}e^{Q_{\theta}(s_t, a)}}
\end{equation}
\end{minipage}
\hfill
\begin{minipage}[t]{0.5\textwidth}
\begin{equation}
\label{equa:2}
J(\theta) = \mathbb{E}_{\tau \sim \pi_\theta} \left[ \sum_{t=0}^{\infty} \gamma^t r_t \right]
\end{equation}
\end{minipage}

where $\gamma \in [0, 1)$ is a discount factor and $\tau$ denotes a trajectory generated by following policy $\pi_\theta$. Optimization proceeds by minimizing a surrogate loss $\mathcal{L}(\theta)$, which may be derived from temporal-difference (TD) errors, policy gradients, or actor-critic estimators \citep{sutton2018}. In TD-based methods, the TD error at timestep $t$ is defined as:
\[
\delta_t = r_t + \gamma V_{\theta}(s_{t+1}) - V_{\theta}(s_t),
\]
where $V_{\theta}(s) = \mathbb{E}_{a\sim\pi_{\theta}(a | s)}Q_{\theta}(s, a)$. The recurrent nature of $\delta_t$ introduces dependencies on both current estimates and future rewards, making $\mathcal{L}(\theta)$ inherently non-stationary. 
As the policy $\pi_{\theta}$ evolves, the data distribution used for training shifts, further complicating optimization. Training is performed by collecting trajectories, computing gradients $\nabla \mathcal{L}(\theta)$, and updating parameters via $\theta \leftarrow \theta - \eta \nabla \mathcal{L}(\theta)$, where $\eta > 0$ is the learning rate. Following conventions from supervised learning, deep RL algorithms often use adaptive variants of stochastic gradient descent, such as Adam~\citep{kingma2014adam} or RMSprop~\citep{hinton2012coursera}, which adjust learning rates based on running estimates of gradient statistics. The gradients with respect to each layer are denoted by;
\[
\nabla \phi_i = \frac{\partial \mathcal{L}}{\partial \phi_i}, \quad \nabla \psi_j = \frac{\partial \mathcal{L}}{\partial \psi_j},
\]
where $\phi_i$ and $\psi_j$ represent the parameters (i.e., weight matrices or bias vectors) of layer $i$ and $j$ respectively.
The structure and magnitude of these gradients ($\nabla \phi_i$ and $\nabla \psi_j$) are influenced by the loss function, data distribution collected from the environment, and the architecture itself. These per-layer gradients determine how effectively different parts of the network adapt during training.

While training large models in supervised learning settings present challenges, advances in initialization, normalization, and scaling strategies have enabled relatively stable optimization \citep{ioffe2015batch,ba2016layer,glorot2010understanding}.
Scaling up model size has been a central driver of progress across domains, improving generalization, enhancing representation learning, and boosting downstream performance \citep{kaplan2020scaling}. 

Deep RL differs substantially from supervised learning. First, the data distribution is non-stationary, continually shifting as $\pi_{\theta}$ updates. Second, learning signals are often sparse, delayed, or noisy, which introduces variance in the estimated gradients~\citep{han2022off,fujimoto2018addressing,liu2025the}. These factors destabilize optimization and lead to loss surfaces with sharp curvature and complex local structure~\citep{Ilyas2020A,achiam2019towards}. Moreover, increasing model capacity often degrades performance unless regularization or architectural interventions are applied~\citep{gogianu2021spectral,bjorck2021towards,schwarzer23bbf,wang20251000}.

These challenges are further compounded by  both architectural and environmental factors. Network depth, width, initialization, and nonlinearity affect how gradients are propagated across layers. 
Meanwhile, reward sparsity, exploration dificulty, and transition stochasticity impose additional structure on the optimization landscape. The resulting geometry reflects the joint dynamics of policy, environment, and architecture, making deep RL optimization uniquely complex.

\paragraph{Gradient Propagation} 

Training deep networks poses fundamental challenges for effective gradient propagation \citep{glorot2010understanding}. As network depth increases, gradients may either vanish or explode as they are backpropagated through multiple layers, impeding the optimization of early layers and destabilizing learning dynamics \citep{ba2016layer}. These issues arise from repeated applications of the chain rule. For a network with intermediate hidden representations $\{h_0, h_1, \dots, h_L\}$, where $h_k \in \mathbb{R}^{d_k}$, the gradient of the loss $\mathcal{L}$ with respect to a hidden layer $h_{\ell}$ is:
\[
\frac{\partial \mathcal{L}}{\partial h_{\ell}} = \left( \prod_{k=\ell+1}^{L} \frac{\partial h_k}{\partial h_{k-1}} \right) \frac{\partial \mathcal{L}}{\partial h_L},
\]
where each $\frac{\partial h_k}{\partial h_{k-1}} \in \mathbb{R}^{d_k \times d_{k-1}}$ is the Jacobian. If the singular values of these Jacobians are not properly controlled, their repeated multiplication can cause the norm of the gradient to shrink or grow exponentially with $\mathcal{L}$. This severely impairs convergence, as earlier layers receive little to no useful gradient signal or become numerically unstable \citep{ioffe2015batch, he2016deep}.

In addition to depth, the width of the network also influences gradient propagation. Consider a fully connected layer with weight matrix $W \in \mathbb{R}^{m \times n}$ and input vector $h \in \mathbb{R}^n$. The output is $Wh \in \mathbb{R}^m$, and under the assumption that $W$ and $h$ have i.i.d. zero-mean entries with finite variance $\sigma_W$ and $\sigma_h$, respectively, the variance of the output is given by $\mathrm{Var}[Wh] = n \, \sigma_W \, \sigma_h$.
Thus, scaling the width $n$ without adjusting $\sigma_W$ and $\sigma_h$ leads to instability in forward and backward signal propagation affecting gradient norms and optimization trajectories.

Beyond depth and width, the choice of nonlinearity also plays a central role in determining how gradients propagate \cite{}. In a typical feedforward network, hidden activations evolve as $h_k = \zeta(W_k h_{k-1})$, where $\zeta(\cdot)$ is a nonlinear activation function (e.g., ReLU, $\tanh$, sigmoid), and $W_k$ is the weight matrix at layer $k$. During backpropagation, the gradient with respect to a hidden layer includes the product of the Jacobian of the linear transformation and the derivative of the nonlinearity:
\[
\frac{\partial \mathcal{L}}{\partial h_{k-1}} = W_k^\top \left( \zeta'(W_k h_{k-1}) \odot \frac{\partial \mathcal{L}}{\partial h_k} \right),
\]
where $\zeta'(\cdot)$ denotes the elementwise derivative of the activation function, and $\odot$ represents elementwise multiplication. For ReLU, $\zeta'(x) = \mathbf{1}_{x > 0}$, so the gradient is entirely blocked wherever the neuron is inactive. This leads to the well-known \emph{dying ReLU} problem, where a significant portion of the network ceases to update and becomes untrainable~\citep{lu2019dying, shin2020trainability}.

\section{Diagnosis: Gradients Under Non-stationarity and Scale}
\label{sec:diagnosis}

A fundamental premise of modern deep learning is that scaling model capacity yields consistent gains in performance~\citep{kaplan2020scaling,chowdhery2023palm}. This has held true in large-scale supervised learning, where training data distributions are stationary and i.i.d., and gradient descent operates under relatively stable conditions. However, in non-stationary settings, such as RL, gradient-based optimization faces severe challenges that scaling alone may exacerbate \citep{ceron2024value,ceron2024mixtures}. In this section, we diagnose how gradient pathologies emerge and intensify across different settings, with a focus on architectural scaling in width and depth (network scales used specified in Table~\ref{tab:mlp_parameters}).

\subsection{Gradient Pathologies}
We train neural networks of varying depths and widths and analyze their training dynamics.

\begin{figure}[!h]
    \centering
    \includegraphics[width=0.30\linewidth]{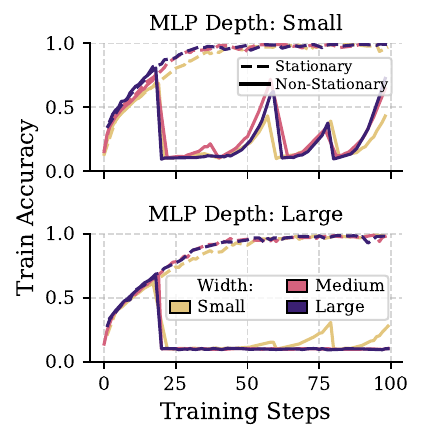}%
    \vline height 122pt depth 0 pt width 1.3 pt
     \includegraphics[width=0.7\linewidth] {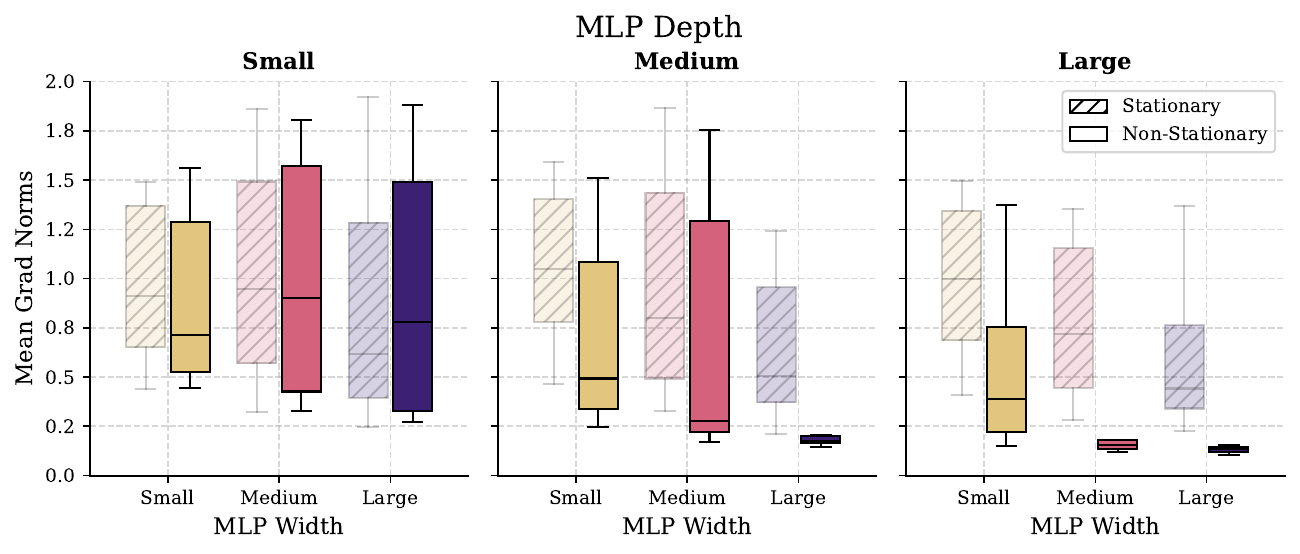}%
    \caption{\textbf{Training dynamics under stationary and non-stationary supervised learning.} (Left) In the stationary setting, both shallow and deep models fit the data effectively across widths. Under non-stationarity only shallow networks partially recover during training, while deeper ones collapse.
    (Right) This collapse correlates with degraded gradient flow. In stationary settings, gradient norms remains stable across all network scales  (\sethlcolor{pastelMediumLight}\hl{shaded boxes}) while in non-stationary settings (\hlMedium{solid-colored boxes}), gradient magnitudes diminish with depth and width, suggesting poor adaptability.}
    \label{fig:supervisedStationary}
\end{figure}

\textbf{Supervised Learning (Stationary and Non-Stationary)}
We use the CIFAR-10 image classification benchmark \citep{krizhevsky2009learning}, where the input-output mapping remains fixed over time. Models consist of standard 6-layer convolutional neural networks (CNN) followed by a multi-layer perceptron (MLP). We vary the depth and width of the MLP to explore how model scale influences learning behavior. To introduce non-stationarity, we periodically shuffle the training labels during training, following the setup by \citet{sokar2023dormant}. This creates a loss landscape that changes over time, echoing the challenges of deep RL. \autoref{fig:supervisedStationary} illustrates the contrast in training behavior and gradient flow between stationary and non-stationary supervised learning. Under non-stationarity, deep networks fail to recover accuracy, which aligns with a marked degradation in gradient magnitudes.

\begin{figure}[!h]
    \centering
    \includegraphics[width=0.3\linewidth]{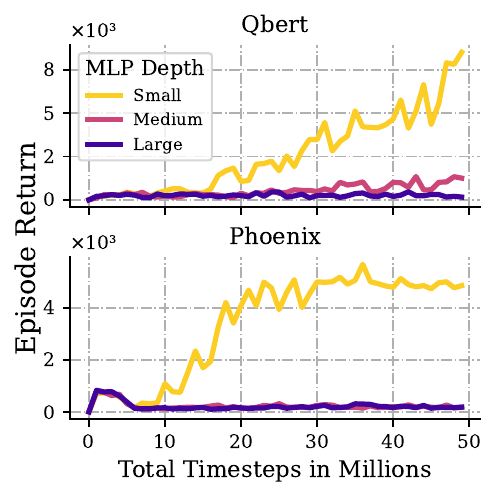}%
    \vline height 122pt depth 0 pt width 1.3 pt
    \includegraphics[width=0.7\linewidth]{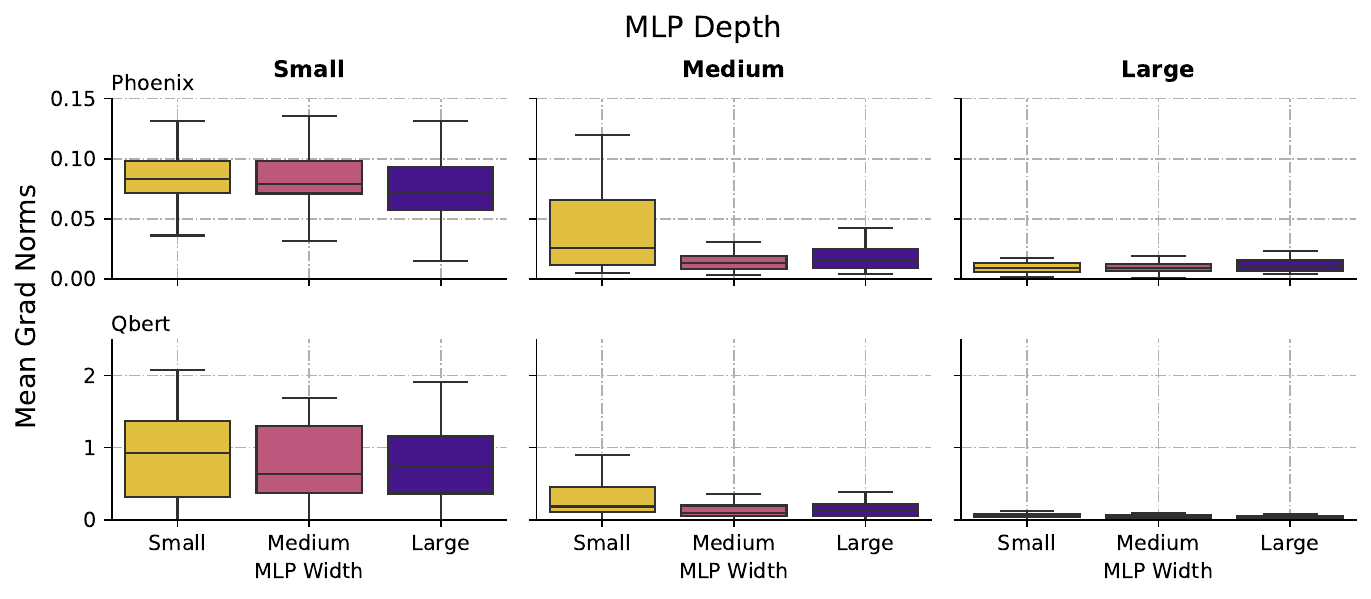}%
    \caption{\textbf{Mean episode returns and gradient norms across increasing MLP depths and widths} on two ALE games using PQN. (Left) Only shallow networks achieve high episode returns; performance collapses for deeper networks. (Right) The collapse correlates with vanishing gradient norms, suggesting that deeper models fail to adapt to non-stationarity in deep RL.}
    \label{fig:deepRL_gradients}
\end{figure}

\textbf{Reinforcement Learning}
As discussed in \autoref{sec:preliminaries}, RL introduces fundamentally different sources of non-stationarity due to the policy-dependent data distribution and moving target estimates. To study gradient dynamics, we use PQN \citep{gallici2025simplifying}, a recent value-based algorithm that achieves strong performance without relying on a target network or replay buffer. PQN ensures stability and convergence using Layer Normalization \citep{ba2016layer} and supports GPU-based training through vectorized environments for online parallel data collection. In \autoref{app:dqn_and_rainbow} we extend our investigation to DQN \citep{mnih2015humanlevel} and Rainbow \citep{hessel2018rainbow}, demonstrating the generality of our observations. As shown in \autoref{fig:deepRL_gradients}, deeper networks trained with PQN exhibit a collapse in both episode returns and gradient norms\footnote{Unless otherwise specified, all ALE results are averaged over three seeds.}, highlighting the fragility of deep models under non-stationarity.

\subsection{Training Degradation}
\label{sec:expressivity-plasticity}
\begin{figure}[!t]
    \centering
    \includegraphics[width=\linewidth]{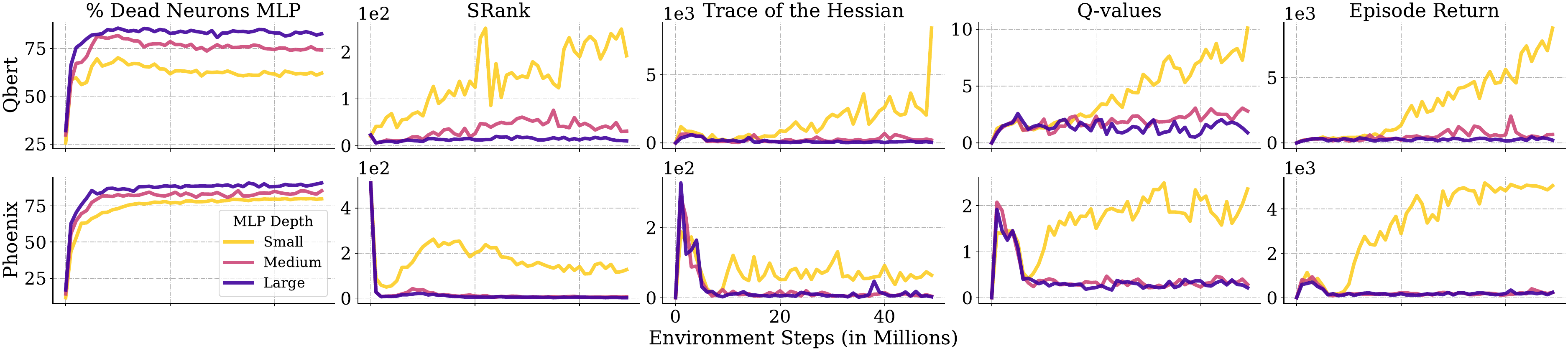}
    \caption{\textbf{Training pathologies emerge as MLP depth increases.} Deeper networks exhibit a higher fraction of inactive neurons, reduced representation rank (SRank), vanishing Hessian trace (loss curvature), and degraded learning performance (mean Q-values and episode returns). These trends indicate that scaling depth limits expressivity and plasticity, impairing policy quality.}
    \label{fig:expressivity}
\end{figure}

In~\autoref{fig:expressivity} we evaluate diagnostic metrics capturing expressivity and training dynamics, revealing that deeper networks exhibit pronounced training pathologies and degraded performance.
We first measure the fraction of \emph{dormant neurons}, defined as units with near-zero activations over a batch of trajectories \citep{sokar2023dormant}, and find that dormant neurons grow with depth, signaling underutilized capacity. Next, we assess representational expressivity using \textit{SRank}, the effective rank of penultimate-layer activations \citep{kumar2020implicit}, observing that deeper networks tend to collapse state representations into lower-dimensional, and less expressive (as evidenced by declining returns) subspaces. To study loss curvature, we compute the Hessian trace of the temporal-difference loss. This metric serves as a proxy for sharpness or smoothness in optimization \citep{ghorbani2019investigation}, similarly to tracking the largest eigenvalue. \autoref{fig:expressivity} shows that only shallow networks exhibit high Hessian trace values, suggesting access to sharper regions of the loss surface with pronounced directions of improvement. In contrast, deeper architectures consistently show near-zero trace, indicating poorly conditioned geometry that hinders effective gradient-based updates. 
These findings suggest a breakdown in representation, plasticity, and optimization as networks scale, ultimately impeding learning.

\begin{tcolorbox}[colback=gray!10,
leftrule=0.4mm,top=0mm,bottom=0mm]
\textbf{Key observations on gradients under non-stationarity and scale:}
\begin{itemize}
    \item Non-stationarity amplifies gradient degradation in deeper and wider networks.
    \item In deep RL, deeper models suffer from vanishing gradients, reduced activations, and loss of representational expressivity.
    \item The flat loss curvature intensifies with depth, correlating with poor learning.
\end{itemize}
\end{tcolorbox}

\section{Stabilizing Gradients}
\label{sec:stabilizingGradients}
Having identified the pathologies that emerge in non-stationary regimes, particularly under large-scale architectures, we investigate strategies to mitigate these instabilities. We focus on two complementary interventions: skip connections \citep{he2016deep} and optimizers \citep{martens2015optimizing}, as these directly improve gradient flow. 
We continue to use PQN as our base RL algorithm and evaluate on the Atari-10 suite \citep{aitchison2023atari}. In \autoref{sec:beyondALE}, we demonstrate that the effectiveness of our proposed gradient interventions generalize beyond this specific algorithm and environment suite.

\subsection{\hlSmall{Intervention 1}: Multi-Skip Residuals for Gradient Stability}
\label{sec:multiskip}
Gradient instability in deep networks is often aggravated by increasing depth, non-linear activations, and misaligned curvature across layers. While standard residual connections offer some relief by introducing shortcut paths for gradient flow \citep{he2016deep}, they typically span only one or two layers, which can be insufficient in the presence of severe gradient disruption due to non-stationarity. We introduce \emph{multi-skip residual connections}, in which the flattened convolutional features are broadcast directly to all subsequent MLP layers. This design ensures that gradients can propagate from any depth back to the shared encoder without obstruction.

\begin{figure}[!h]
     \includegraphics[width=0.24\textwidth]{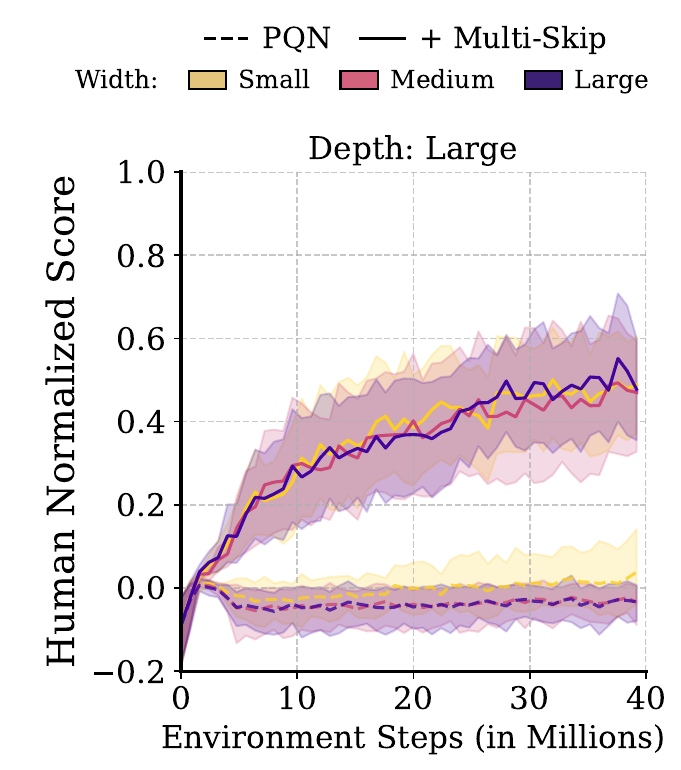}%
     \includegraphics[width=0.24\textwidth]{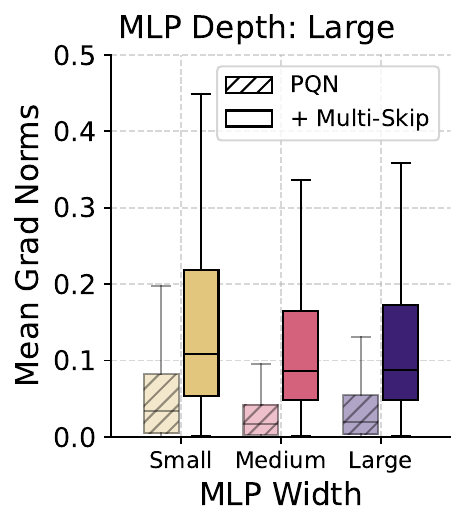}%
    \vline height 110pt depth 0 pt width 1.3 pt
    \includegraphics[width=0.24\textwidth]{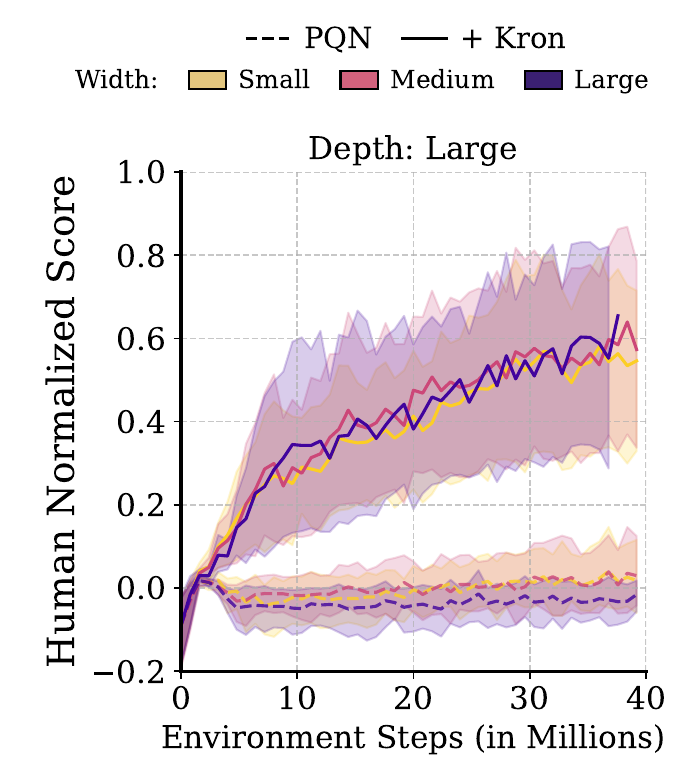}%
     \includegraphics[width=0.24\textwidth]{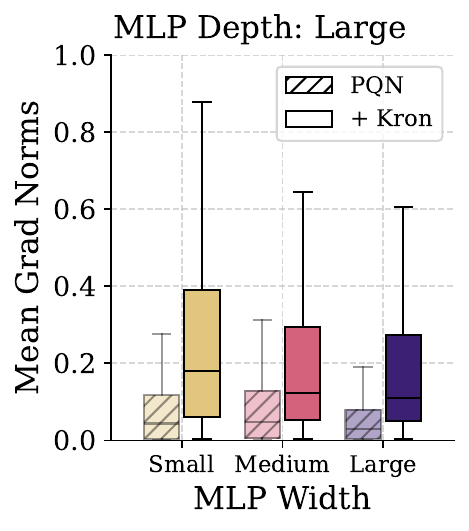}%
    \caption{\textbf{Gradient-stabilizing interventions improve scalability in deep RL.} (Left) Standard fully connected networks trained with PQN collapse at greater depths due to vanishing gradients. In contrast, multi-skip architectures maintain gradient flow and scale effectively.
    (Right) The default RAdam optimizer leads to instability in deep networks, while switching to the Kron optimizer  preserves gradient signal and enables stable learning without architectural changes.
    }
    \label{fig:multiskip_and_kron}
\end{figure}

We compare our network architecture against the standard fully connected baseline across varying depths. As shown in \autoref{fig:multiskip_and_kron} (left), performance collapses with increased depth in the baseline, while the multi-skip architecture maintains stable learning and continues to improve across widths. This improvement is accompanied by consistently higher gradient magnitudes.
Complete results across all network depths and widths are presented in \autoref{app:full_multiskip}.

\begin{figure}[!t]
    \centering
    \includegraphics[width=0.8\textwidth]{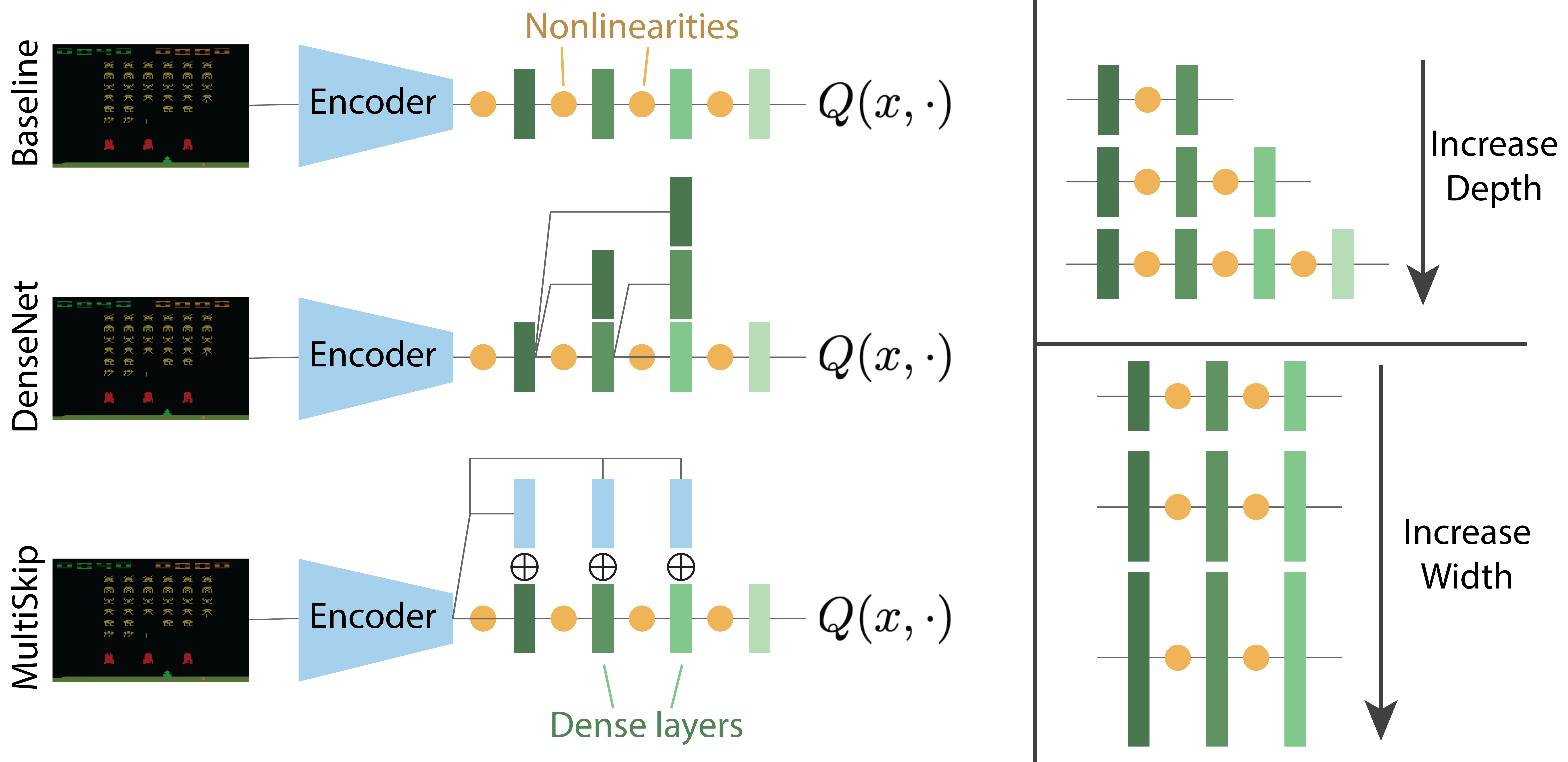}
    \caption{\textbf{(Left)} MLP architectures and \textbf{(right)} scaling strategies studied.}
    \label{fig:architectures}
    \vspace{-0.4cm}
\end{figure}

\vspace{-0.5em}
\subsection{\hlSmall{Intervention 2:} Second-Order Optimizers for Non-Stationarity}
\label{sec:kron}

First-order optimizers such as SGD and Adam rely on local gradient estimates and fixed heuristics (e.g., momentum, adaptive step sizes) \citep{kingma2014adam}, which are agnostic to curvature and often brittle under shifting data distributions. In contrast, second-order methods adjust parameter updates using curvature information, enabling more informed and stable adaptation. 

Let $\mathcal{L}(\theta)$ denote the loss function, and $g = \nabla \mathcal{L}(\theta)$ its gradient. A second-order update takes the form $\theta_{t+1} = \theta_t - \eta H^{-1} g$, where $H$ is the curvature matrix, typically the Hessian or the Fisher Information Matrix (FIM) \citep{martens2020new}. Directly inverting $H$ is computationally infeasible in deep neural networks so Kronecker-factored approximations, such as K-FAC \citep{martens2015optimizing}, address this challenge by approximating $H$ using low-rank Kronecker products.

Kronecker-factored optimizer (Kron for short) approximates the FIM and applies structured preconditioning that captures inter-parameter dependencies, unlike Adam’s diagonal scaling.
This yields directionally aware preconditioning that better aligns with the curvature of the loss surface \citep{martens2020new}. In non-stationary settings, such as deep RL, where both the data distribution and curvature evolve over time, curvature-aware updates can help preserve gradient signal by maintaining stable update magnitudes and directions. 
As shown in \autoref{fig:multiskip_and_kron} (right), replacing RAdam with Kron prevents performance collapse at greater depths, even in standard MLP architectures.
Complete results across all network depths and widths are presented in \autoref{app:full_kron}.

\subsection{\hlSmall{Combining Gradient Interventions}}
\label{sec:puttingItAllTogether}
\begin{figure}[!h]
    \centering
    \includegraphics[width=0.19\linewidth]{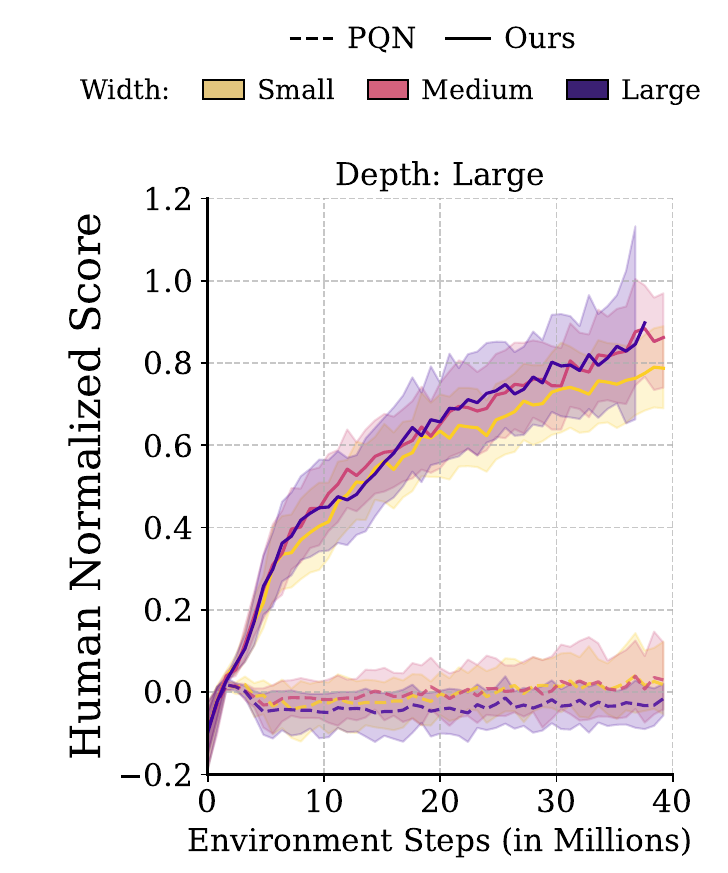}%
    \includegraphics[width=0.19\textwidth]{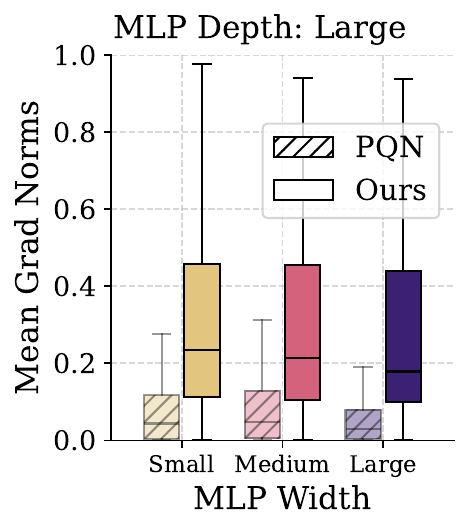}%
       \vline height 90pt depth 0 pt width 1.3 pt
    \includegraphics[width=0.6\linewidth]{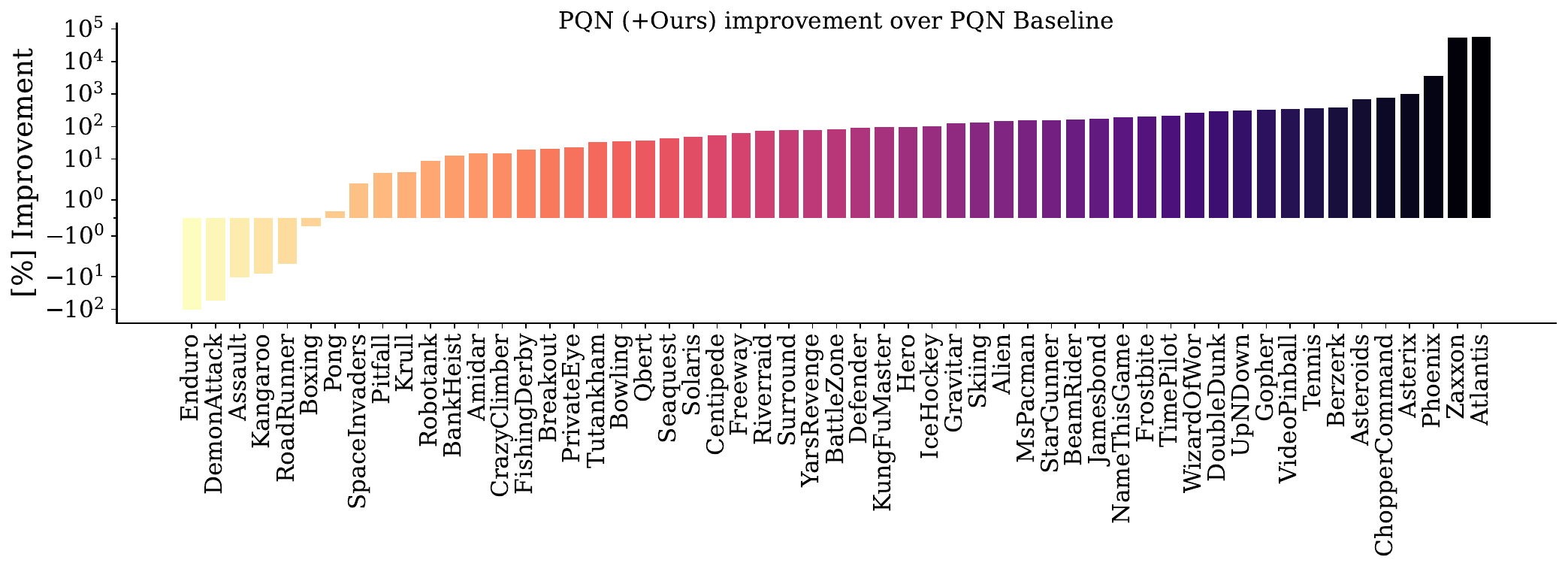}%
    \caption{\textbf{Gradient-stabilized PQN achieves superior scalability.} (Left) On Atari-10, the combined interventions lead to high HNS even at greater depths, outperforming either intervention alone (see \autoref{fig:multiskip_and_kron}) and increased gradient gradient flow.
    (Right) On the full ALE suite, our agent outperforms the baseline in 90\% of the games with a median performance improvement of 83.27\%.}
    \label{fig:pqn_combined}
\end{figure}

We combine both gradient interventions to PQN and evaluate it on the full ALE suite (57 games), across 3 seeds and 200M frames.
\autoref{fig:pqn_combined} shows that our augmented agent outperforms the baseline in 90\% of the environments, achieving a median relative improvement of 83.27\%. Notably, the baseline PQN is itself competitive 
with strong agents such as Rainbow \citep{gallici2025simplifying}, highlighting the effectiveness of our interventions. Detailed per-environment learning curves can be found in \autoref{app:fullALE}.

\begin{figure}[!h]
      \includegraphics[width=0.36\linewidth]{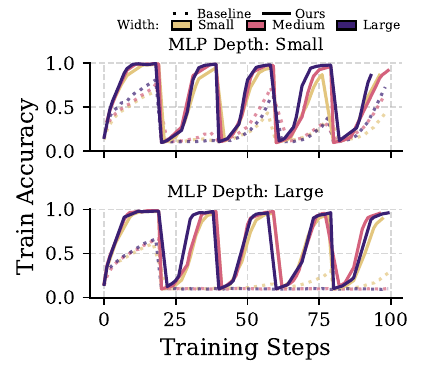}
       \vline height 122pt depth 0 pt width 1.3 pt
\includegraphics[width=0.64\linewidth]{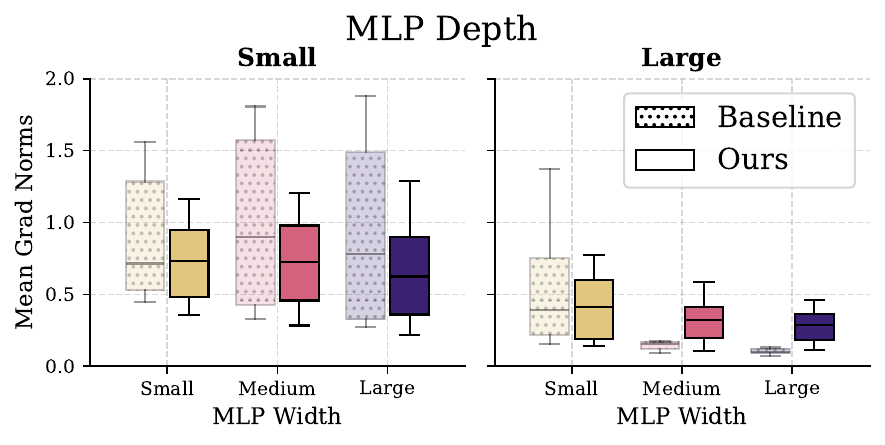}
    \caption{\textbf{Gradient interventions enable rapid recovery in non-stationary SL.} (Left) Models with combined gradient interventions rapidly recover accuracy after label reshuffling, demonstrating robust adaptation in non-stationary settings. (Right) This is supported by stable gradient flow across depth. Dashed curves and shaded boxes indicate MLP baselines.
    }
\label{fig:allTogether_nonstationary_SLMainBody}
    \vspace{-0.2cm}

\end{figure}

In \autoref{fig:allTogether_nonstationary_SLMainBody} we validate the effectiveness of the combined gradient interventions in the non-stationary SL setting we used as motivation in \autoref{sec:diagnosis}.  The results verify that these interventions enable high accuracy and sustained adaptability across depths and widths, even under dynamic label reshuffling.

\subsection{Alternative Gradient-Stabilization Methods}
\label{sec:other_interventions}
To ensure that our findings were not specific to a narrow choice of interventions, we conducted a broader exploration of alternative strategies for improving gradient stability in deep RL. We tested a variety of approaches inspired by prior work on optimization and representation stability in both supervised and RL settings (see \autoref{app:additional_interventions} for more details). 

As summarized in \autoref{tab:alternative_interventions}, none of these methods consistently improved stability or performance compared to our proposed combination of multi-skip residuals and Kronecker-factored optimization. In many cases, the alternatives yielded either negligible gains or degraded performance as network depth increased, reinforcing that architectural and curvature-aware interventions are key to preserving gradient flow at scale.
\section{Beyond the ALE and PQN}
\label{sec:beyondALE}

To evaluate the generality of our findings, we extend our analyses. Specifically, we: \textit{(i)} apply our proposed methods to PPO~\citep{schulman2017proximal} on the full ALE and on continuous control tasks in Isaac Gym~\citep{makoviychuk2021isaac}; \textit{(ii)} assess the impact of richer convolutional encoders by replacing the standard CNN backbone used in the ALE with the Impala CNN architecture~\citep{espeholt2018impala}; \textit{(iii)} augment Simba~\citep{lee2025simba} with our proposed techniques and evaluate performance on the DeepMind Control Suite (DMC)~\citep{tassa2018deepmind}; and \textit{(iv)} investigate whether our interventions can stabilize and scale a wide range of Q-learning algorithms in challenging offline RL benchmarks~\citep{park2025horizon}.

\begin{figure}[!h]
    \includegraphics[width=0.65\linewidth]{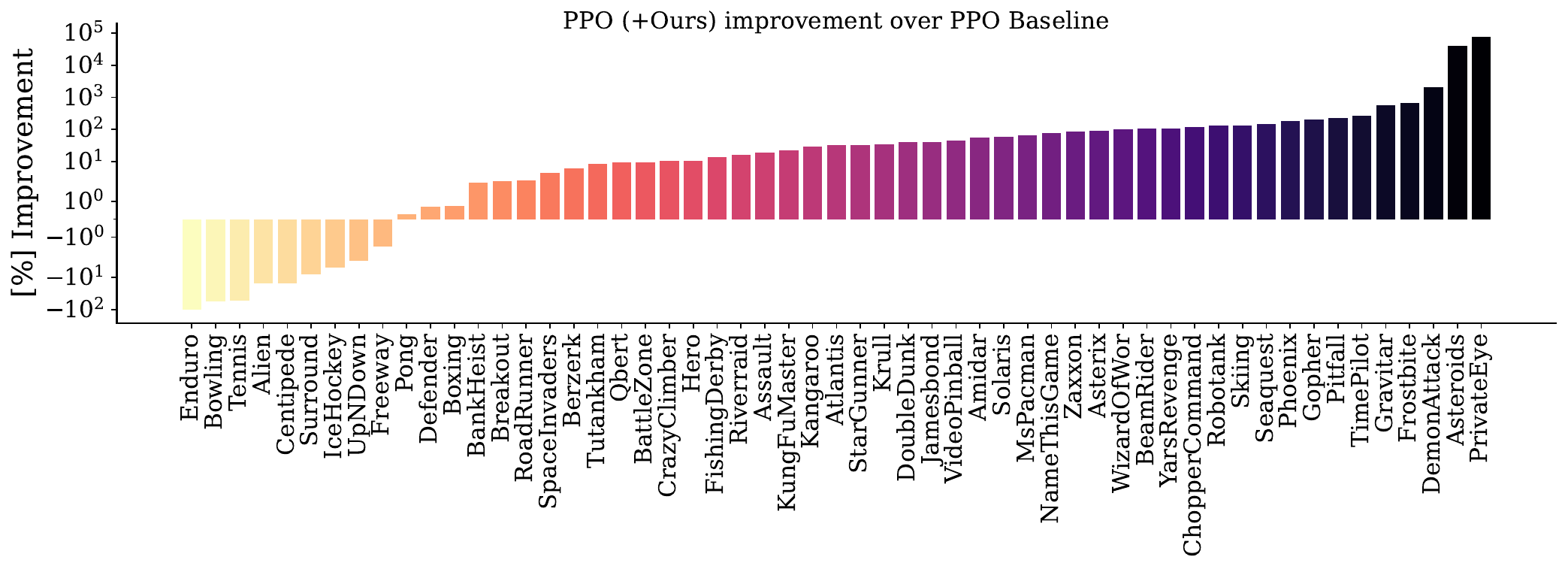}
       \vline height 110pt depth 0 pt width 1.3 pt
       \includegraphics[width=0.35\linewidth]{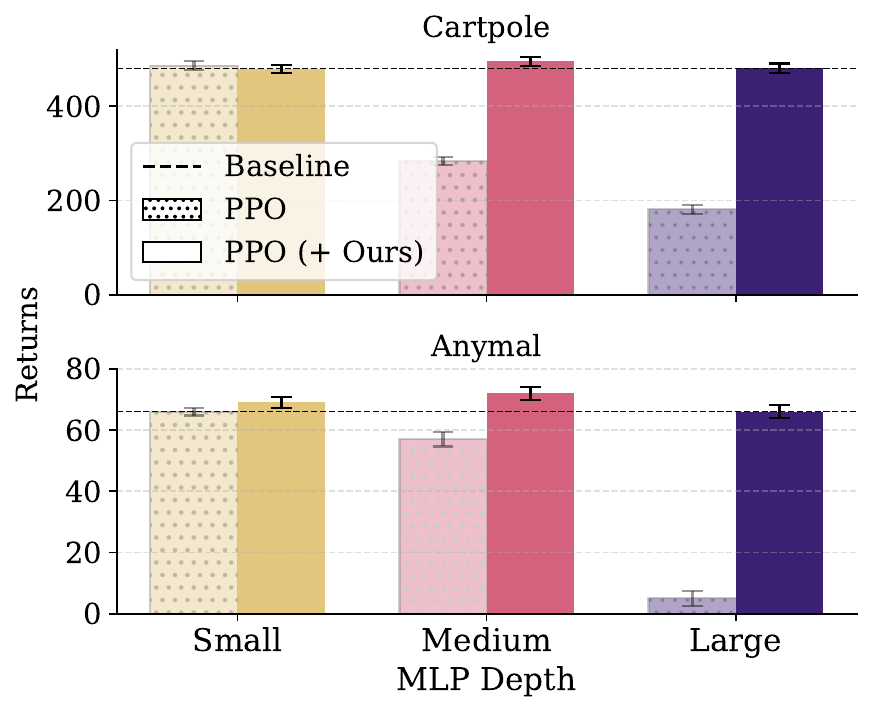}%
    \vspace{-0.2cm}
    \caption{\textbf{PPO with gradient interventions.} Left: On the full ALE suite, applying the combined gradient interventions to PPO yields a median performance improvement of 31.40\% and outperforms the baseline in 83.64\% of the games. Right: In the Cartpole and Anymal tasks from IsaacGym, only the augmented PPO maintains stable performance across depths and widths.}
    \label{fig:ppo_allTogether}
\end{figure}

\paragraph{PPO with Gradient Interventions.} \autoref{fig:ppo_allTogether} (left) shows that augmenting PPO with the same strategies as in PQN (Layer Normalization by default on PQN, multi-skip residual connections, and Kronecker-factored optimization) significantly boots performance. On the ALE benchmark, the augmented PPO outperforms the baseline in 83.64\% of the environments, achieving a median relative improvement of 31.40\%. In Isaac Gym's continuous control tasks, including Cartpole and Anymal (\autoref{fig:ppo_allTogether}, right), the baseline PPO collapses as model size increases, while the augmented variant remains stable and achieves superior performance at all depths and widths.

\begin{figure}[!h]
    \centering
    \includegraphics[width=0.495\linewidth]{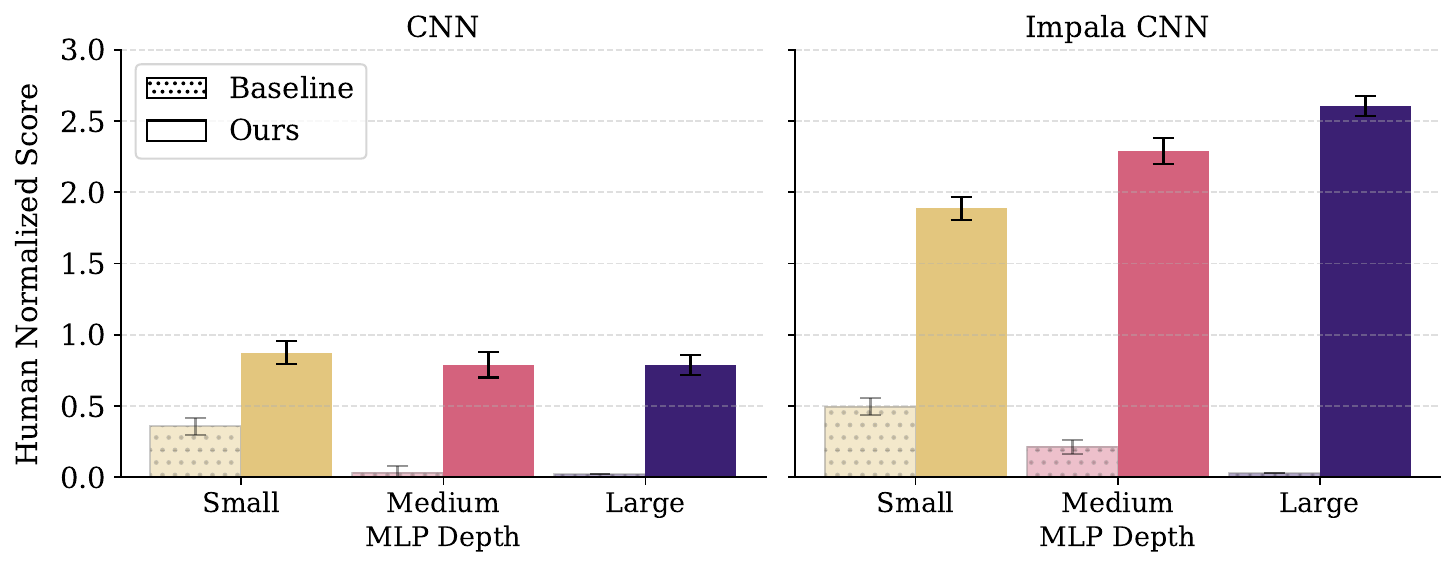}%
    \vline height 85pt depth 0 pt width 1.3 pt
    \includegraphics[width=0.495\linewidth]{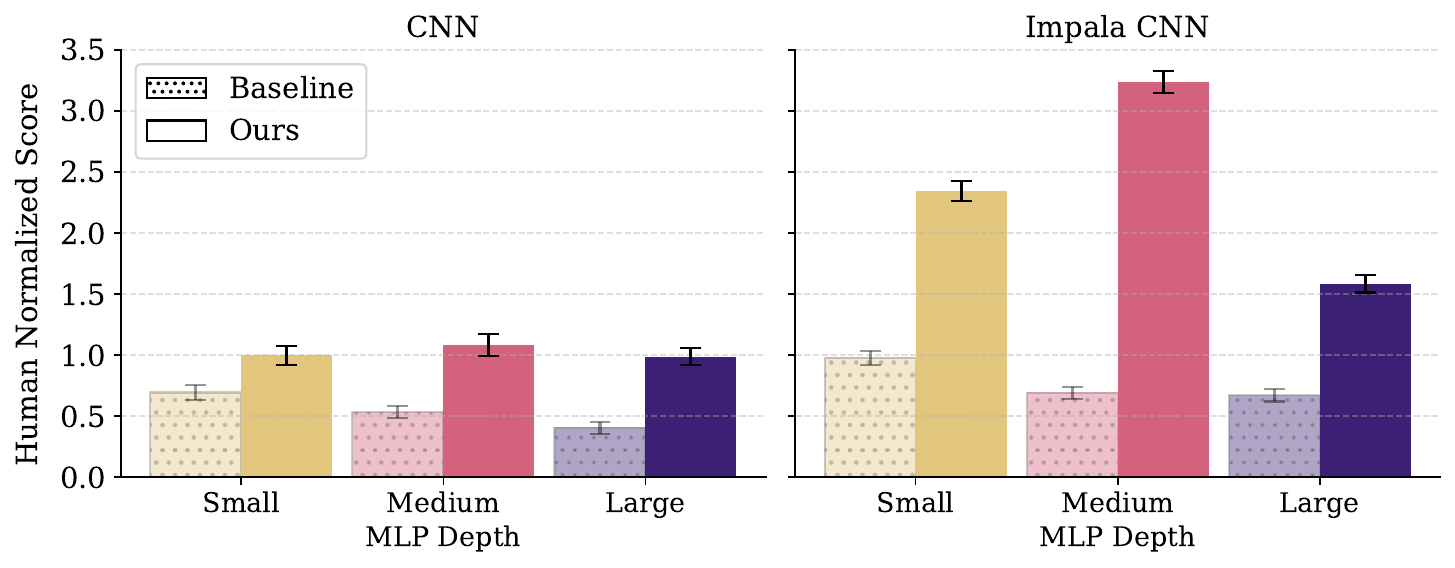}%
    \caption{\textbf{Scaling performance with standard vs. Impala CNN encoders} on PQN (left) and PPO (right). Each agent is evaluated using both the Atari CNN (left sub-panels) and the Impala CNN (right sub-panels) as the encoder. Gradient interventions enable successful scaling in both cases.
    }
    \label{fig:ppo_and_pqn_impalacnn}
\end{figure}

\paragraph{Gradient Interventions in Scaled Encoder Variants} 
The Impala CNN is a scalable convolutional architecture that has demonstrated strong performance gains in agents such as Impala~\citep{espeholt2018impala} and Rainbow~\citep{hessel2018rainbow}. We investigate whether, given its capacity to extract richer representations from visual input, combining Impala CNN with our gradient flow interventions enables effective scaling of the MLP component. As shown in \autoref{fig:ppo_and_pqn_impalacnn}, PPO and PQN benefit significantly from replacing the standard CNN with the Impala CNN. For PQN, the Impala encoder enables successful scaling of the MLP, in contrast to the performance collapse seen without our interventions. These results suggest that the expressivity of richer visual encoders is more effectively leveraged by deeper networks when gradient flow is preserved. 

\begin{figure}[!t]
    \centering
    \includegraphics[width=\linewidth]{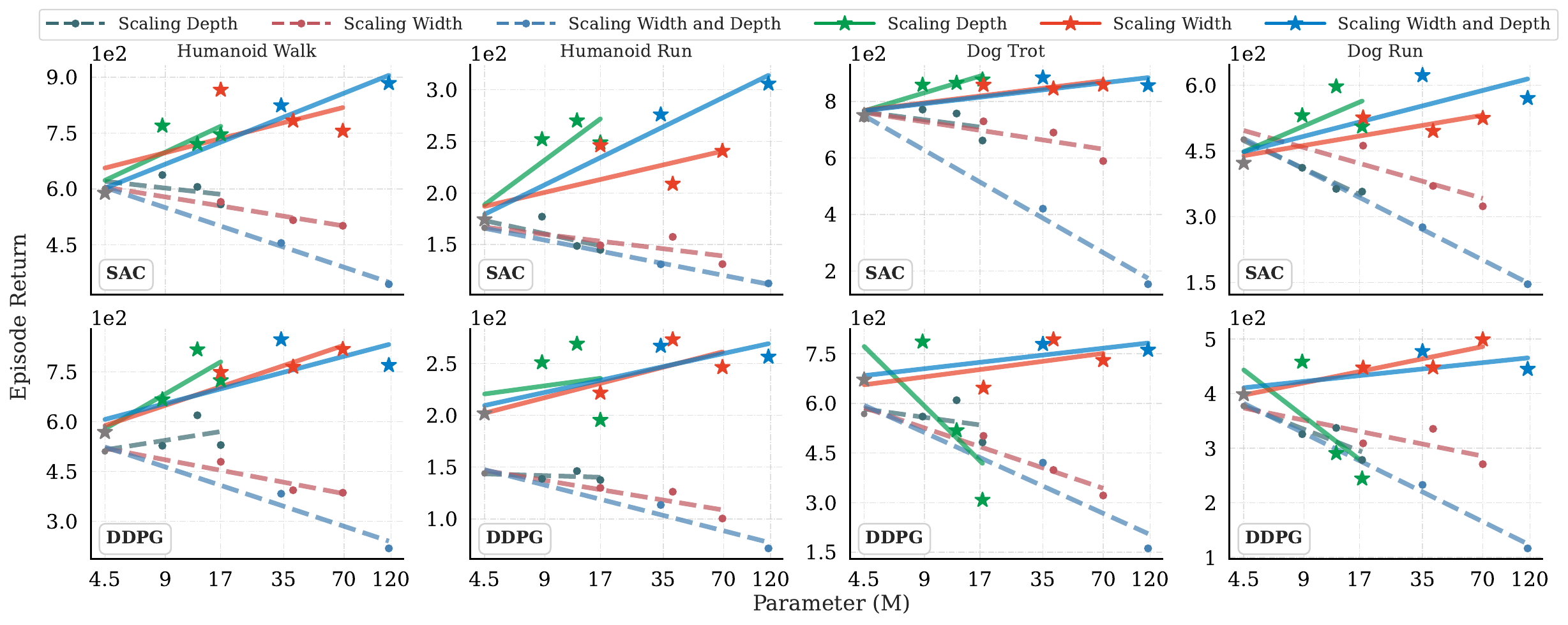}
    \vspace{-0.7cm}
    \caption{\textbf{Performance comparison between AdamW (dashed lines) and Kron (solid lines) optimizers using the SimBa architecture with SAC and DDPG,} averaged over $5$ random seeds. As model size increases, AdamW leads to consistent performance degradation, while Kron enables stable and improved learning with larger networks.}
    \label{fig:DMC-all}
    \vspace{-0.3cm}
\end{figure}

\vspace{-1.5em}
\paragraph{Simba with Kron Optimizer.} Simba \citep{lee2025simba} is a scalable actor-critic framework that integrates normalization, residual connections, and LayerNorm. We augment Simba by replacing its default AdamW optimizer with Kron while keeping all other hyperparameters fixed. We evaluate SAC~\citep{haarnoja2018soft} and DDPG~\citep{lillicrap2015continuous} on challenging DMC tasks, using the Simba architectures of varying depth and width.
Despite its design for scalability, default Simba collapses across all tasks as networks grow as shown in \autoref{fig:DMC-all} (additional results in \autoref{sec:simbaFullResults}). In contrast, the Kron-augmented version successfully scales in both depth and width, achieving consistent and stable performance gains. These findings underscore the generality of our approach as effectively enabling parameter scaling in deep RL agents.

\begin{figure}[!h]
    \centering
    \includegraphics[width=\linewidth]{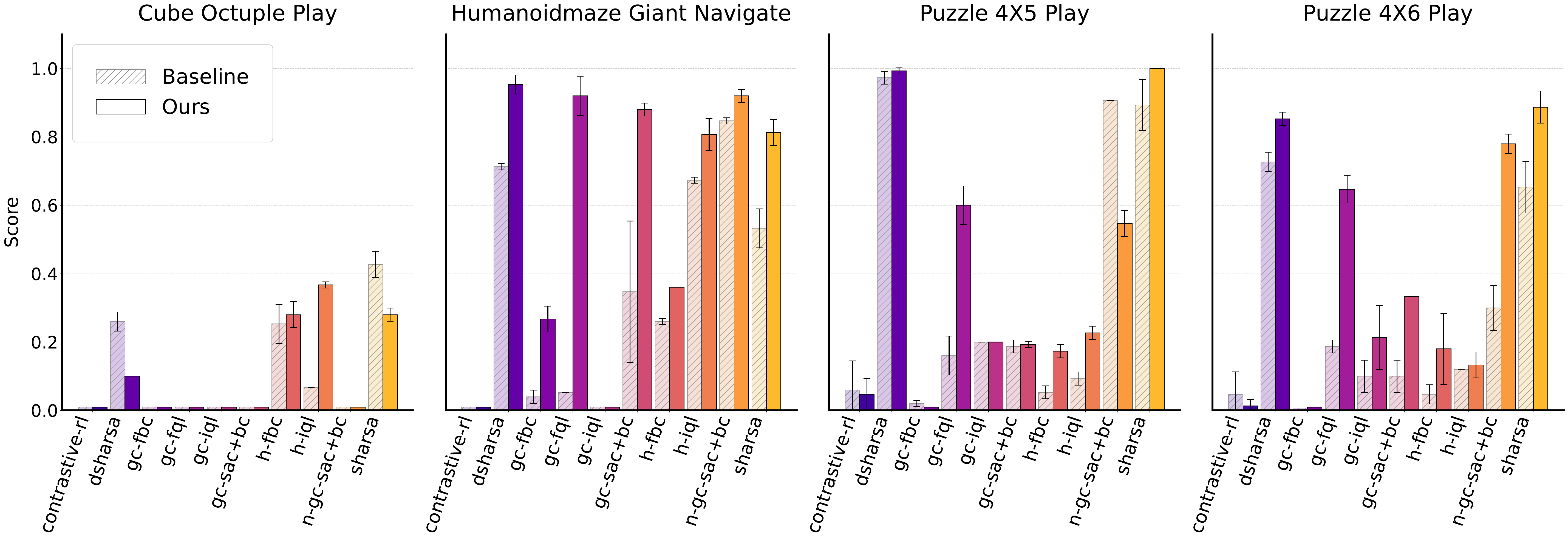}
    \vspace{-0.6cm}
    \caption{\textbf{Performance of offline Q-learning algorithms with and without our gradient interventions.} We compare the original algorithms from \cite{park2025horizon} against our augmented versions. The results, averaged over $3$ seeds demonstrate a general improvement in scalability.}
    \label{fig:scalable_q_offline}
    \vspace{-0.3cm}
\end{figure}

\paragraph{Gradient Interventions for Scalable Offline Q-Learning.}
\citet{park2025horizon} highlight significant challenges in scaling Q-learning algorithms for complex offline RL tasks, demonstrating that many standard offline RL baselines fail to learn effective policies, even on large, high-quality datasets. Their key finding was that performance improvements were primarily driven by techniques that shorten the effective credit assignment horizon, such as n-step returns and hierarchical methods. This led to their proposal of two new high-performing algorithms, SHARSA and DSHARSA, which are designed to operate with shorter effective horizons.  This finding motivates a parallel investigation: can our proposed gradient interventions, which are designed to stabilize and accelerate deep network training, also address the scaling limitations of offline Q-learning? To test this, we augment the full suite of baselines and novel algorithms from \cite{park2025horizon} with our proposed gradient interventions. The results, presented in \autoref{fig:scalable_q_offline}, show that our methods provide a complementary path to scalability. Applying our interventions generally improves the performance of the baselines across all tasks. 

The performance gains are particularly pronounced in the most sparse-reward task, \texttt{humanoidmaze-giant-navigate}, where our gradient interventions enable multiple methods to achieve near-optimal performance, whereas their baseline counterparts largely fail. Furthermore, this stability extends to generalization across task difficulty. When moving from \texttt{puzzle-4x5-play} to the harder \texttt{puzzle-4x6-play} task, many baselines exhibit a sharp performance degradation. In contrast, the performance of several algorithms with our interventions remains consistent and high, demonstrating improved robustness. Finally, we note that while the primary focus of this paper is to address gradient pathologies arising from scaling and non-stationarity, these results highlight that our interventions are also highly beneficial in offline deep RL, where inputs are stationary.

\section{Related Work}
\label{sec:related_work}

A central challenge in scaling deep RL lies in the inefficient use of model capacity. Increasing parameter counts often fails to yield proportional gains due to under-utilization. \citet{sokar2023dormant,mayor2025the} and \citet{liu2025measure} show that online RL induces a growing fraction of inactive neurons, a phenomenon also observed in offline settings. \citet{ceron2024value} report that up to 95\% of parameters can be pruned post-training with negligible performance drop, underscoring substantial redundancy. These findings have motivated techniques such as weight resetting~\citep{schwarzer23bbf}, tokenized computation~\citep{sokar2025dont}, and sparse architectures~\citep{ceron2024mixtures,willi2024mixture,liu2025neuroplastic,ma2025network}, along with auxiliary objectives to promote capacity utilization~\citep{farebrother2023protovalue}. While scaling model size offers greater expressivity, its benefits depend on appropriate training strategies~\citep{ota2021training}. Architectural interventions such as SimBa~\citep{lee2025simba} improve robustness by regularizing signal propagation through components such as observation normalization, residual feedforward blocks, and layer normalization. Complementarily, BRO~\citep{naumanbigger} shows that scaling the critic network yields substantial gains in sample and compute efficiency, provided it is paired with strong regularization and optimistic exploration strategies.

Gradient flow, however, remains a central bottleneck. We complement prior efforts by explicitly targeting vanishing gradients as a mechanism for improving scalability. Our approach builds on the role of LayerNorm in stabilizing training and enhancing plasticity~\citep{lyle2024normalization}, and leverages its theoretical effect on gradient preservation as formalized in PQN~\citep{gallici2025simplifying}. Optimization-level interventions such as second-order methods~\citep{martens2015optimizing, muppidi2024fast} and adaptive optimizers~\citep{ellis2024adam, bengio2021correcting, wu2017scalable} also address instability under non-stationarity. Our approach integrates architectural and optimizer-level interventions to enable stable gradient flow and unlock parameter scaling in deep RL agents.
\section{Discussion}
\label{sec:discussion}

Our analyses in \autoref{sec:diagnosis} suggest that the difficulty in scaling networks in deep RL stems from the interaction between inherent non-stationarity and gradient pathologies that worsen with network size. In \autoref{sec:stabilizingGradients}, we introduced targeted interventions to address these challenges, and in \autoref{sec:puttingItAllTogether}, we demonstrated their effectiveness.
We validated the generality of our approach across agents and environment suites, consistently observing similar trends. These findings reaffirm the critical role of network design and optimization dynamics in training scalable RL agents. While our proposed solutions may not be optimal, they establish a strong baseline and provide a foundation for future work on gradient stabilization in deep RL.
More broadly, our findings suggest that scaling limitations in deep RL are not solely attributable to algorithmic instability or insufficient exploration, but also stem from gradient pathologies amplified by architectural and optimization choices. Addressing these issues directly, without altering the learning algorithm, yields substantial gains in scalability and performance. This suggests that ensuring stable gradient flow is a necessary precondition for effective parameter scaling in deep RL. \\

{\bf Limitations.} Our study is constrained by computational resources, which limited our ability to explore architectures beyond a certain size. While our interventions show consistent improvements across agents and environments, further scaling remains an open question. While using second order optimizers introduced additional computational overhead (see \autoref{tab:exp_times}), this cost is mitigated by leveraging vectorized environments and efficient deep RL algorithms, narrowing the gap relative to standard methods. These limitations highlight promising directions for future work, including the development of more computationally efficient gradient stabilization strategies and scalable optimization techniques.

\newpage

\paragraph{Acknowledgment}
The authors would like to thank João Guilherme Madeira Araújo, Evan Walters, Olya Mastikhina, Dhruv Sreenivas, Ali Saheb Pasand, Ayoub Echchahed and Gandharv Patil for valuable discussions during the preparation of this work. João Araújo deserves a special mention for providing us valuable feed-back on an early draft of the paper. We want to acknowledge funding support from Natural Sciences and Engineering Research Council (NSERC) of Canada, Google Research, Fonds de recherche du Québec (FRQNT) and The Canadian Institute for Advanced Research (CIFAR) and compute support from Digital Research Alliance of Canada, Mila IDT, and NVidia. We would also like to thank the Python community \cite{van1995python, 4160250} for developing tools that enabled this work, including NumPy \cite{harris2020array}, Matplotlib \cite{hunter2007matplotlib}, Jupyter \cite{2016ppap}, and Pandas \cite{McKinney2013Python}.

\paragraph{Broader Impact} This paper presents work whose goal is to advance the field of Machine Learning. There are many potential societal consequences of our work, none which we feel must be specifically highlighted here.

\bibliographystyle{plainnat}
\bibliography{references}

\clearpage
\appendix
\section{Environment Details}
\label{app:environment_details}
Throughout the paper, we evaluate the deep reinforcement learning agents' performance on the Atari-10 suite \citep{aitchison2023atari}, a curated subset of games from the Arcade Learning Environment (ALE) \citep{bellemare2013arcade}. Atari-10 consists of 10 games selected to capture the maximum variance in algorithm performance, achieving over 90\% correlation with results on the full ALE benchmark. This makes it a computationally efficient yet representative testbed for deep reinforcement learning. We follow the experimental protocol of \citet{obando2023small, ceron2024mixtures, agarwal2021deep}, running each experiment with three random seeds and reporting the aggregate human-normalized score across games.

The games in Atari-10 are:

\begin{itemize}
\item Amidar, Battle Zone, Bowling, Double Dunk, Frostbite, Kung Fu Master, Name This Game, Phoenix, Q*Bert and River Raid.
\end{itemize}

Additionally, to further support the generality of our findings, we evaluate the proposed combined gradient interventions on the full ALE benchmark. We also assess their effectiveness on continuous control tasks from the IsaacGym simulator \citep{makoviychuk2021isaac} and the DeepMind Control Suite (DMC) \citep{tassa2018deepmind}, extending our analysis to robotics-based environments. We conduct experiments on the 4 challenging tasks of DMC:
\begin{itemize}
\item
Humanoid Walk, Humanoid Run, Dog Trot and Dog Run.
\end{itemize}

\section{Network Sizes}
Throughout the paper, we experiment with models of varying depths and widths. Unless stated otherwise (e.g. in ~\autoref{sec:beyondALE}, where we evaluate the Impala CNN), the convolutional feature extractors are kept fixed. Consequently, our experiments focus primarily on scaling strategies and architectural variations in the MLP components of the networks.

To enable meaningful comparisons across different learning regimes, the MLP architectures are kept consistent across supervised learning (SL), non-stationary SL, and reinforcement learning (RL) experiments. This consistency ensures that observed differences in gradient behavior arise from the learning setting itself, rather than confounding factors due to domain-specific architectures.

Table~\ref{tab:mlp_parameters} provides detailed information on the number of parameters for each depth–width configuration, categorized as small, medium, or large, as used throughout the paper.

\begin{table}[!h]
\centering
\caption{Number of parameters (in millions) for different MLP architectures.}
\label{tab:mlp_parameters}
\begin{tabular}{lcccc}
\toprule
\textbf{Depth / Width} & \textbf{\cellcolor{WidthSmall!25}Small} & \textbf{\cellcolor{WidthMedium!25}Medium} & \textbf{\cellcolor{WidthLarge!25}Large} & \\
\midrule
\textbf{Small}  & \cellcolor{WidthSmall!10}2.39 & \cellcolor{WidthMedium!10}11.90 & \cellcolor{WidthLarge!10}27.70 &  \\
\textbf{Medium} & \cellcolor{WidthSmall!10}3.45 & \cellcolor{WidthMedium!10}21.35 & \cellcolor{WidthLarge!10}53.93 &  \\
\textbf{Large}  & \cellcolor{WidthSmall!10}4.50 & \cellcolor{WidthMedium!10}30.79 & \cellcolor{WidthLarge!10}80.15 &  \\
\bottomrule
\end{tabular}
\end{table}

\section{Additional Experiments}
\subsection{Scaling with DQN and Rainbow}
\label{app:dqn_and_rainbow}

To further support our hypothesis on the emergence of gradient pathologies in deep reinforcement learning, we investigate whether similar issues arise in algorithms beyond PQN and PPO, as discussed in the main paper. Specifically, we study the effects of architectural scaling on two widely used value-based algorithms: DQN \citep{mnih2015humanlevel} and Rainbow \citep{hessel2018rainbow}.

DQN is a foundational deep RL algorithm that learns action-value functions using temporal difference updates and experience replay, serving as a standard baseline for value-based methods. Rainbow extends DQN by integrating several enhancements, such as double Q-learning, prioritized experience replay, dueling networks, multi-step learning, distributional value functions, and noisy exploration, to achieve improved sample efficiency and stability.

\begin{figure}[!h]
    \centering
    \begin{subfigure}[t]{0.49\linewidth}
        \centering
        \includegraphics[width=\linewidth]{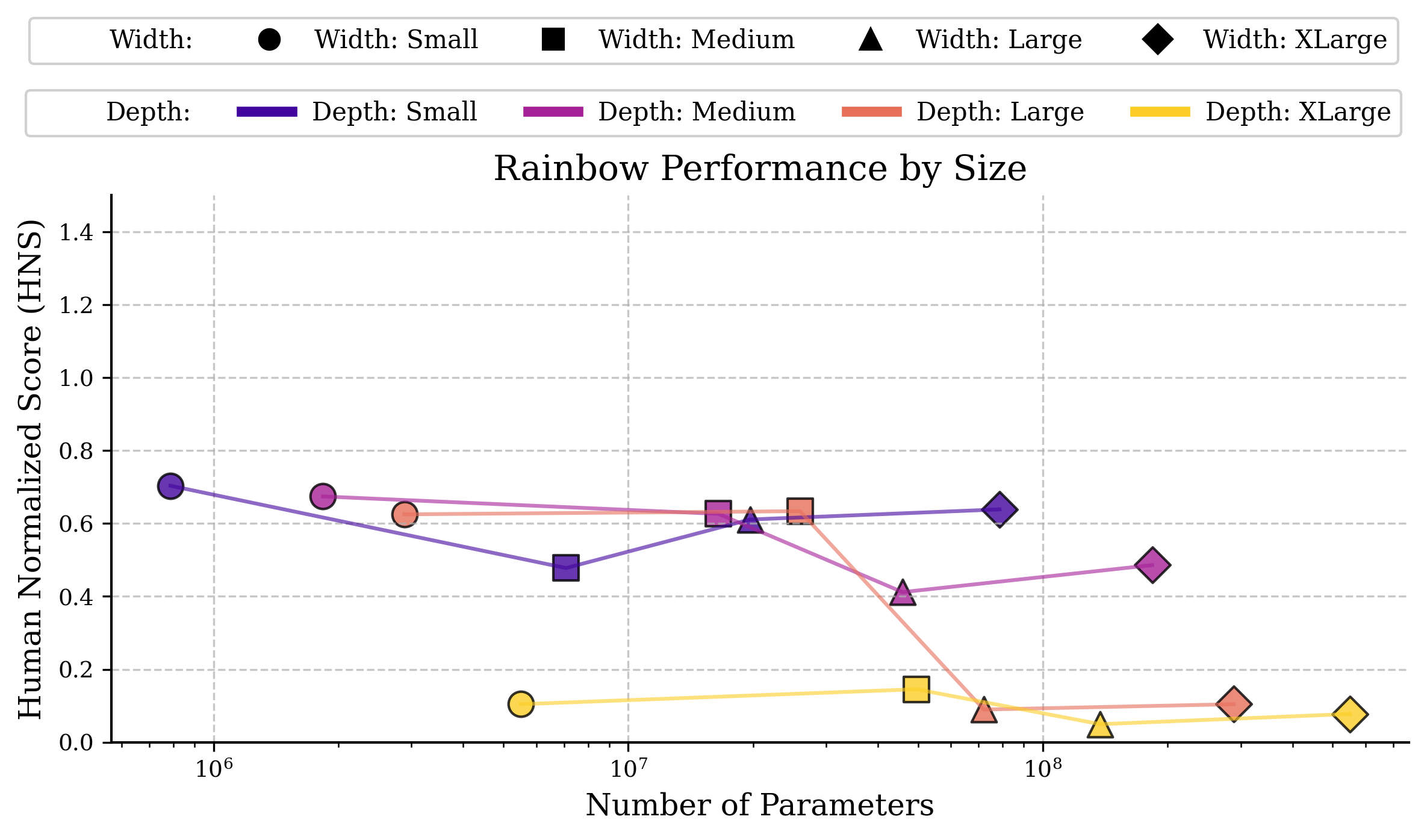}
        \caption{DQN HNS}
        \label{fig:dqn_hns}
    \end{subfigure}
    \hfill
    \begin{subfigure}[t]{0.49\linewidth}
        \centering
        \includegraphics[width=\linewidth]{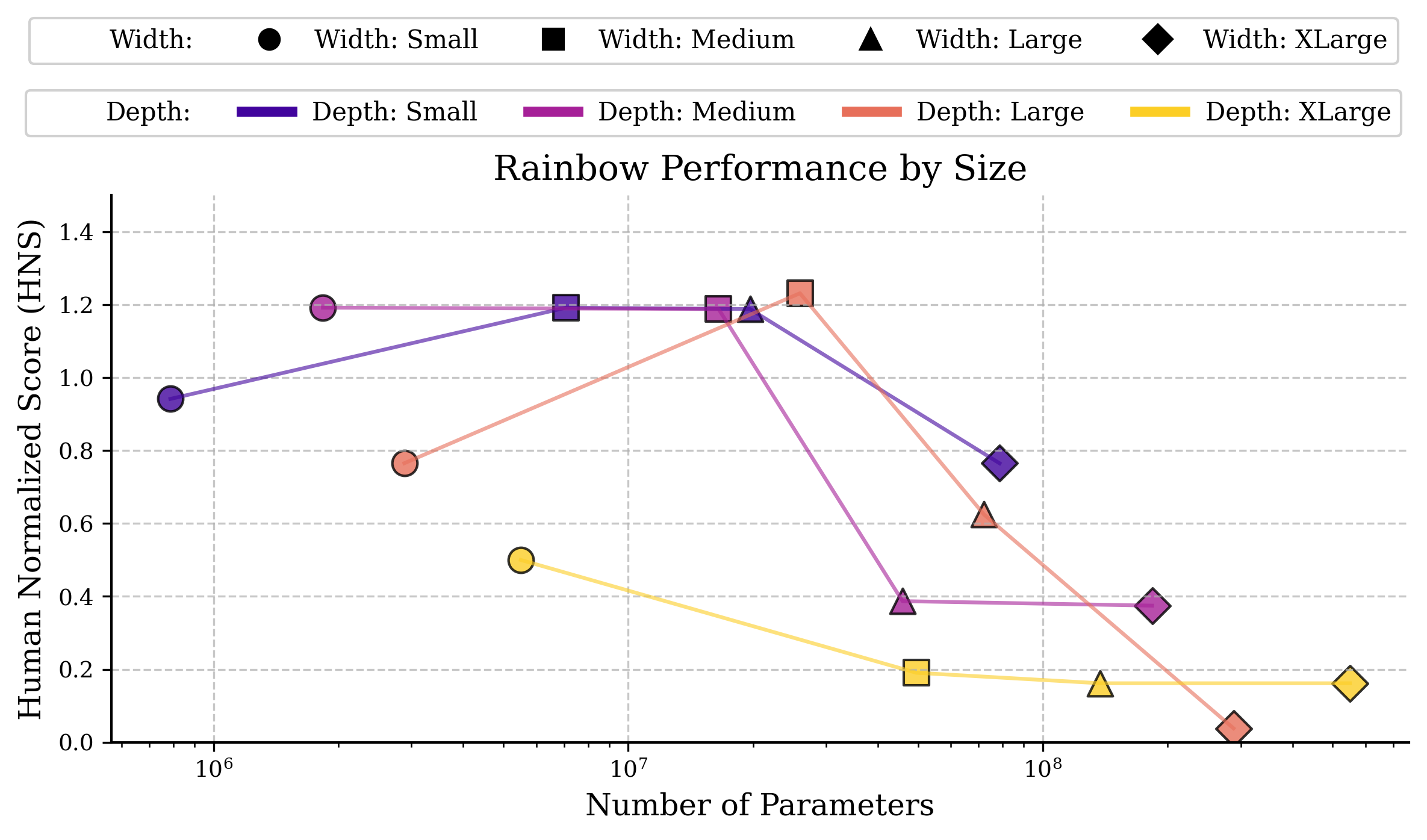}
        \caption{Rainbow HNS}
        \label{fig:rainbow_hns}
    \end{subfigure}
    \caption{\textbf{Median human normalized scores for DQN (left) and Rainbow (right) as a function of total network parameters.} Lines of different colors denote varying network depths, while marker shapes indicate different widths. For both agents, performance consistently declines as network size increases, highlighting the adverse effects of scaling.}
    \label{fig:dqn_and_rainbow_hns}
\end{figure}

\begin{figure}[!h]
    \centering
    \includegraphics[width=\linewidth]{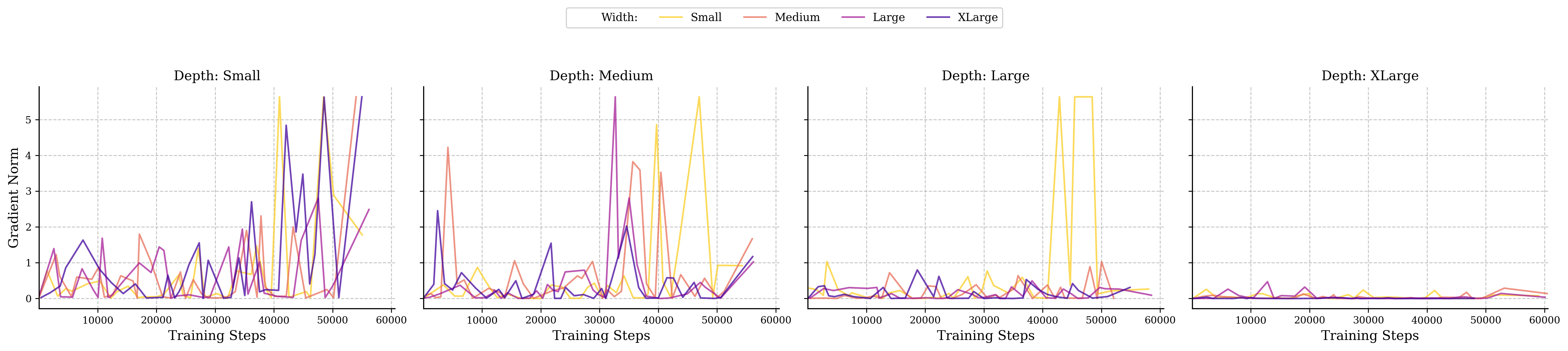}
    \includegraphics[width=\linewidth]{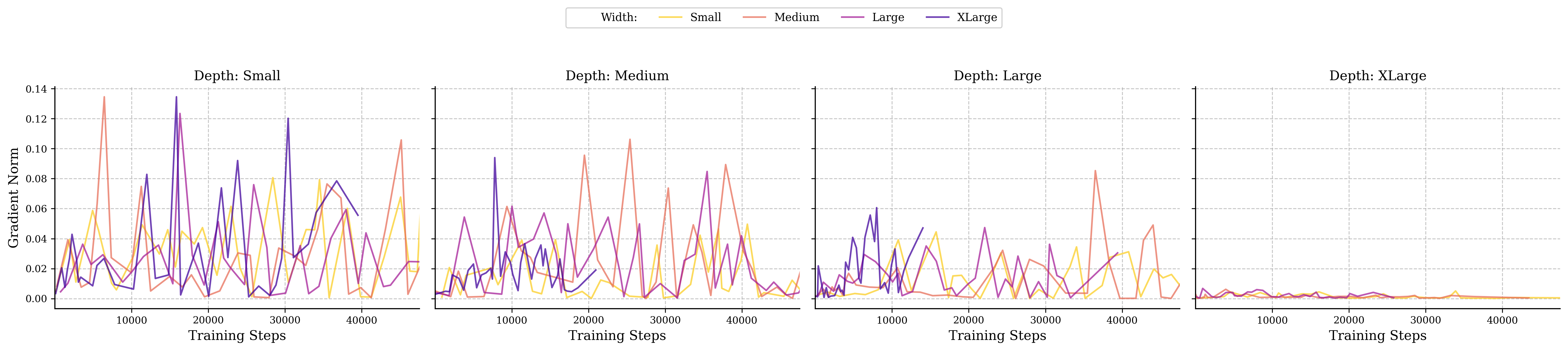}
    \caption{\textbf{Gradient magnitudes during training for DQN (top) and Rainbow (bottom).} As network depth increases, gradient flow systematically diminishes, ultimately collapsing to near-zero values. This consistent decay mirrors the performance degradation observed at larger scales.}
    \label{fig:dqn_and_rainbow_gradients}
\end{figure}

In \autoref{fig:dqn_and_rainbow_hns}, we report the performance of DQN and Rainbow as we scale the depth and width of their networks. As with PQN and PPO, we observe consistent degradation in performance at larger scales. In \autoref{fig:dqn_and_rainbow_gradients}, we present the corresponding gradient behavior, which reveals the same vanishing and destabilization phenomena discussed in this work. These findings reinforce the generality of the identified gradient pathologies across both policy-based and value-based deep RL algorithms.

\subsection{Combining Gradient Interventions in Non-stationary Supervised Learning}

Building on our findings in \autoref{sec:puttingItAllTogether}, we extend our analysis by applying the proposed combined gradient interventions to the same image classification models used in \autoref{sec:diagnosis}. Specifically, we train the models in the non-stationary supervised learning setup, where the CIFAR-10 labels are iteratively shuffled, following the experimental design from \citet{sokar2023dormant}. As demonstrated in \autoref{sec:diagnosis}, while models in standard supervised learning settings are able to scale effectively and maintain high performance, introducing non-stationarity leads to failure in adaptation for baselines that use fully connected layers and the Adam optimizer. This issue is exacerbated as model scale increases.

Our results, presented in \autoref{fig:allTogether_nonstationary_SLMainBody}, show that combining the multi-skip architecture for the MLP component with the Kronecker-factored optimizer and Layer Normalization enables near-perfect continuous adaptation. The models quickly adapt to the changing optimization problem following label reshuffling, with gradient magnitudes remaining stable throughout the process.

\subsection{Architecture and Optimizer Ablations}
\label{app:heatmaps}

In this work, we introduce the multi-skip architecture, an extension of the standard residual MLP design, and propose the use of the Kronecker-factored optimizer for online deep RL. While these techniques form the basis of our primary interventions, our broader goal is not to prescribe a fixed set of methods, but rather to motivate a general class of architectural and optimization interventions that promote healthy gradient flow in deep networks. To this end, we expand the scope of our evaluation by incorporating a wider range of baselines. Specifically, we compare various optimizer choices, including Adam and AdaBelief \citep{zhuang2020adabelief}, alongside MLP architectures such as the standard residuapl MLP \citep{he2016deep} and DenseNet \citep{huang2017densely}. These architectures have been previously explored in the context of scaling networks in online deep RL \citep{lee2025simba, ota2024framework}, providing a relevant basis for comparison.

We also evaluated state-of-the-art optimizers that have demonstrated success in training large-scale models such as transformers in supervised learning. Specifically, we tested Shampoo \citep{gupta2018shampoo}, a second-order optimizer that maintains and preconditions gradients using full-matrix statistics per layer, and Apollo \citep{ma2020apollo}, an adaptive optimizer that leverages curvature information without explicitly computing or storing second-order matrices.

Despite extensive hyperparameter tuning for both methods, we were unable to achieve strong performance in the online deep RL setting. This suggests that further investigation is needed to understand the key properties required for these optimizers to be effective in this regime. \citet{asadi2023resetting,ceron2021revisiting} demonstrate that optimizer behavior plays a critical role in the training dynamics of online deep RL methods, with \citet{asadi2023resetting} showing that stale optimizer states can hinder learning, and \citet{ceron2021revisiting} revealing that optimizer sensitivity interacts with the choice of loss function, particularly when comparing Huber and MSE losses.

\begin{figure}[!h]
\includegraphics[width=\linewidth]{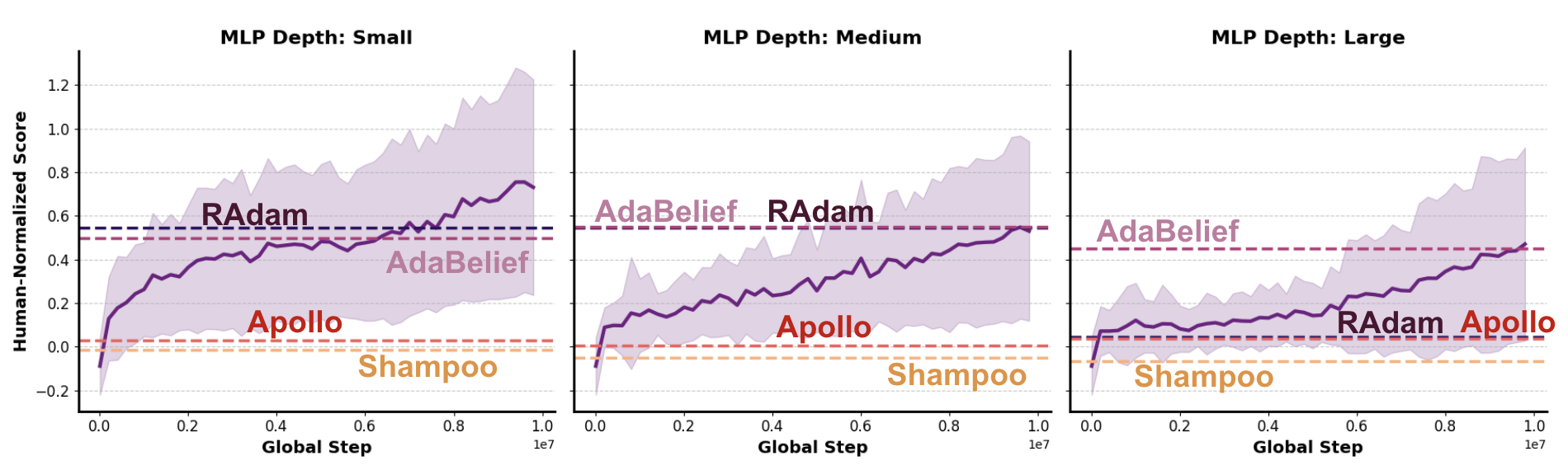}
\includegraphics[width=\linewidth]{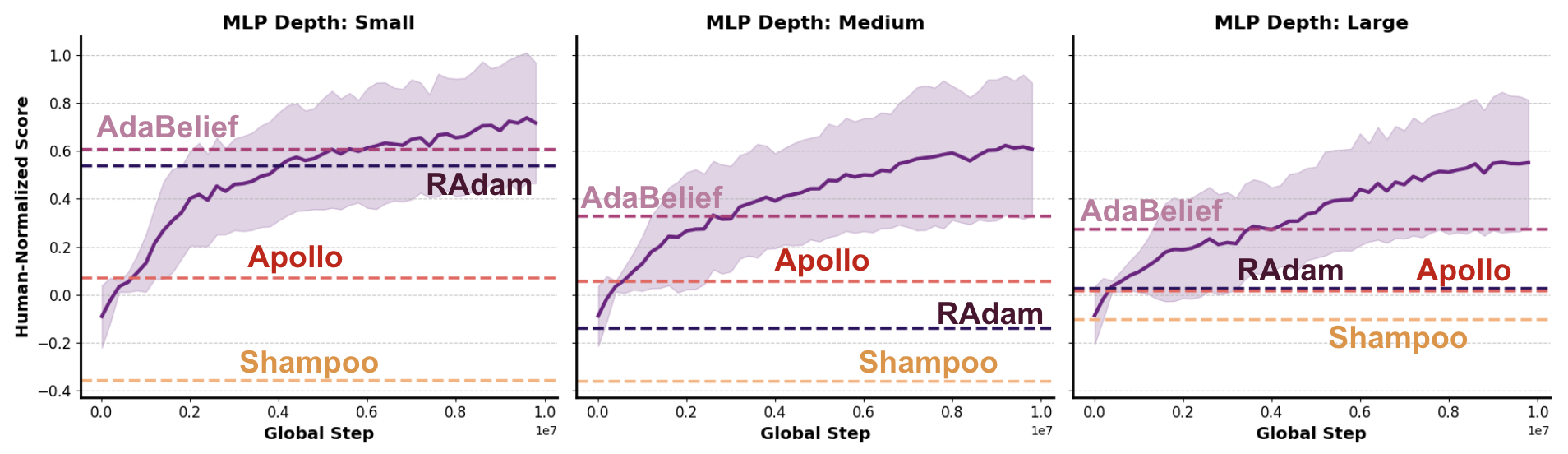}
\caption{\textbf{Median human normalized scores on Atari-10 for PPO (top row) and PQN (bottom row), comparing a range of optimizers including RAdam, AdaBelief, Shampoo, Apollo, and Kron (shown in the main curves).} While adaptive optimizers like AdaBelief show some robustness, only Kron consistently enables stable and performant training as models scale. Each curve represents the mean performance across three random seeds
per algorithm, with shaded areas indicating 95\% bootstrap confidence intervals.}
\label{fig:optimizer_baselines}
\end{figure}

We present the results for PPO and PQN across all tested optimizers in \autoref{fig:optimizer_baselines}.

\subsection{Results with the Multi-Skip Architecture.}
\label{app:full_multiskip}

We present the full learning curves comparing the proposed multi-skip architecture to the baseline fully connected architecture across all depths and widths studied in the paper. We follow the experimental protocol of \citet{obando2023small, ceron2024mixtures, agarwal2021deep}, running each experiment with three random seeds.

\begin{figure}[!h]
\includegraphics[width=\linewidth]{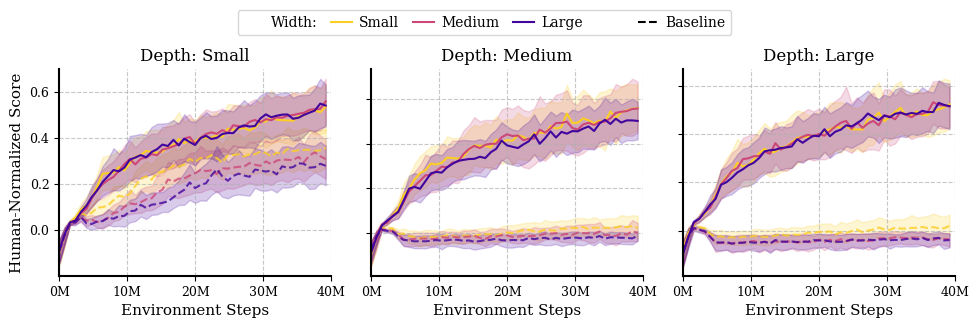}
\caption{\textbf{Median human-normalized scores with PQN on the Atari-10 benchmark, comparing the baseline agent and the proposed multi-skip architecture across varying depths and widths.} The multi-skip architecture not only improves performance at shallow depths, but also enables PQN to remain trainable across all scales considered, whereas the baseline MLP rapidly collapses as depth and width increase. Each curve represents the mean performance across three random seeds
per algorithm, with shaded areas indicating 95\% bootstrap confidence intervals.}
\label{fig:full_multiskip}
\end{figure}

\subsection{Ablation on the number of skip connections.}

To isolate the effect of skip length on performance, we fix the main network of our proposed MultiSkip architecture (in \textit{large} size, which includes 5 residual blocks) and vary how many of these blocks receive skip connections from the encoder. When Skip = $k$, we apply the encoder features as skip connections to the first $k$ residual blocks immediately following the encoder, while the remaining $(5 - k)$ blocks operate without direct encoder input. The table below reports human-normalized scores on the Atari-10 benchmark \cite{agarwal2021deep}.

\begin{table}[h!]
\centering
\caption{Human-normalized scores on the Atari-10 benchmark, varying the number of residual blocks ($k$) that receive skip connections from the encoder. Performance generally improves as more connections are added.}
\label{tab:skip_ablation}
\begin{tabular}{lrrrrr}
\toprule
\textbf{Environment} & \textbf{Skip=1} & \textbf{Skip=2} & \textbf{Skip=3} & \textbf{Skip=4} & \textbf{Skip=5} \\
\midrule
Amidar-v5         & 0.20  & 0.17  & 0.19  & 0.20          & \textbf{0.36}   \\
BattleZone-v5     & 0.01  & 0.67  & 0.62  & 0.60          & \textbf{0.69}   \\
Bowling-v5        & 0.07  & 0.04  & 0.04  & 0.08          & \textbf{0.23}   \\
DoubleDunk-v5     & -2.09 & -2.00 & -1.55 & -1.36         & \textbf{-1.32}  \\
Frostbite-v5      & 0.67  & 0.70  & 0.79  & \textbf{0.92} & 0.88            \\
KungFuMaster-v5   & 0.93  & 0.95  & 0.93  & \textbf{1.12} & 0.98            \\
NameThisGame-v5   & 0.79  & 0.65  & 0.72  & 0.85          & \textbf{1.24}   \\
Phoenix-v5        & \textbf{0.69}  & 0.68  & 0.68  & 0.66          & 0.66            \\
Qbert-v5          & 0.84  & 1.01  & \textbf{1.06}  & 0.91          & 1.09            \\
Riverraid-v5      & 0.42  & 0.44  & 0.65  & 0.69          & \textbf{0.99}   \\
\midrule
\textbf{Aggregate (mean)} & 0.25  & 0.33  & 0.41  & 0.47          & \textbf{0.58}   \\
\bottomrule
\end{tabular}
\end{table}

Performance steadily improves as more skip connections are added, peaking when all 5 blocks are connected. This supports our original design decision to broadcast features to all MultiSkip blocks.

\subsection{Results with the Kron Optimizer.}
\label{app:full_kron}

We present the full learning curves comparing the Kron optimizer to the baseline RAdam optimizer originally used in PQN \citep{gallici2025simplifying}, across all depths and widths studied in the paper. We follow the experimental protocol of \citet{obando2023small, ceron2024mixtures, agarwal2021deep}, running each experiment with three random seeds.

\begin{figure}[!h]
\includegraphics[width=\linewidth]{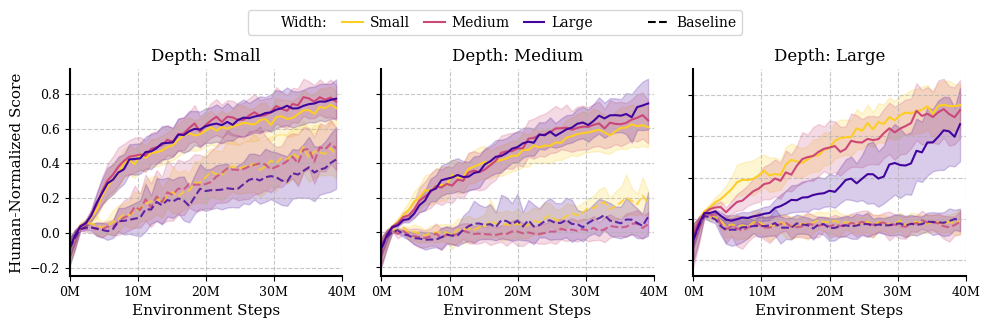}
\caption{\textbf{Median human-normalized scores with PQN on the Atari-10 benchmark, comparing the Kron optimizer to the baseline RAdam optimizer across varying depths and widths.} Similar to the multi-skip architecture, Kron not only improves performance at shallow depths, but also enables PQN to remain trainable across all scales considered. In contrast, performance with RAdam rapidly collapses as depth and width increase. Each curve represents the mean performance across thee random seeds
per algorithm, with shaded areas indicating 95\% bootstrap confidence intervals.}
\label{fig:full_kron}
\end{figure}

\subsection{Justification for Selected Interventions.}
\label{app:additional_interventions}

The choice of Kronecker-factored optimization and the MultiSkip architecture was the result of a systematic exploration of candidate interventions aimed at improving gradient flow, plasticity, and stability in deep RL. We evaluated a wide range of alternative methods from prior work known to mitigate optimization pathologies \citep{moalla2024no,juliani2024a}.

Our evaluated methods, summarized in \autoref{tab:alternative_interventions}, included:
\begin{itemize}
    \item Second-order and adaptive optimizers: Apollo \citep{ma2020apollo}, Shampoo \citep{gupta2018shampoo}, AdaBelief \citep{zhuang2020adabelief}.
    \item Regularization: L2 norm penalties \citep{kumar2023offline}, weight clipping, and weight decay \citep{elsayed2024weight}.
    \item Activation functions: GELU \citep{hendrycks2016gaussian} and CReLU \citep{abbas2023loss}.
    \item Learning rate schedules: Cosine annealing and cyclic schedulers.
    \item Learning rate scaling: Multiplying and dividing the default learning rate (2.5e-4) by 10 to compensate for increased network scale.
\end{itemize}

As shown in the table, none of these interventions consistently improved performance compared to our proposed combination. This motivated our decision to focus on the combination of Kronecker-factored optimization and the multi-skip architecture.

\begin{table}[h!]
\centering
\caption{\textbf{Comparison of mean human-normalized scores on Atari-10 for alternative interventions.} Our proposed method (Ours) is compared against ablations using only a single alternative intervention. The results (3 seeds per experiment) show that no single alternative consistently matches the performance of our combined approach.}
\label{tab:alternative_interventions}
\resizebox{\textwidth}{!}{%
\begin{tabular}{lrrrrrrrrrr}
\toprule
\textbf{Environment} & \textbf{Ours} & \textbf{Cosine LR} & \textbf{Cyclic LR} & \textbf{GELU} & \textbf{CReLU} & \textbf{L2 Norm} & \textbf{Weight Clip} & \textbf{Weight Decay} & \textbf{LR=2.08e-5} & \textbf{LR=3.00e-3} \\
\midrule
Amidar-v5       & \textbf{0.355897} & -0.001663 & 0.002276 & 0.015143 & 0.016514 & 0.029293 & 0.012283 & 0.013013 & 0.018848 & 0.078660 \\
BattleZone-v5   & \textbf{0.694566} & 0.031297 & -0.000287 & 0.015505 & -0.004594 & -0.004594 & 0.035604 & 0.061446 & 0.019812 & 0.575407 \\
Bowling-v5      & \textbf{0.231831} & 0.050145 & -0.008721 & 0.028706 & -0.009811 & -0.000727 & 0.030523 & 0.051235 & -0.086483 & 0.067587 \\
DoubleDunk-v5   & \textbf{-1.318182} & -2.318182 & -2.318182 & -2.363636 & -2.318182 & -2.363636 & -2.454545 & -2.454545 & -2.454545 & -2.090909 \\
Frostbite-v5    & \textbf{0.881087} & -0.001921 & 0.027708 & 0.008890 & -0.010938 & 0.004755 & 0.075372 & 0.021736 & 0.008034 & 0.110036 \\
KungFuMaster-v5 & 0.975251 & -0.011055 & 0.003181 & -0.000823 & -0.011500 & -0.011278 & -0.011500 & -0.011055 & -0.011055 & \textbf{1.362077} \\
NameThisGame-v5 & \textbf{1.242066} & -0.030191 & -0.132767 & -0.261400 & -0.198430 & -0.068147 & -0.197995 & -0.148227 & -0.134417 & 0.747251 \\
Phoenix-v5      & \textbf{0.655141} & 0.072687 & -0.087237 & -0.090014 & -0.090168 & -0.068722 & -0.090940 & -0.093949 & -0.102358 & 0.411968 \\
Qbert-v5        & \textbf{1.094894} & 0.013814 & 0.010052 & -0.002833 & 0.008359 & 0.006102 & 0.016259 & 0.000177 & 0.001305 & 0.112282 \\
Riverraid-v5    & \textbf{0.993568} & -0.000127 & 0.345100 & 0.304636 & -0.056846 & -0.058494 & 0.293514 & -0.044456 & 0.057765 & 0.237365 \\
\bottomrule
\end{tabular}
}
\end{table}

\newpage
\section{Hyper-parameters}
\label{app:hyperparameters}

Below, we report the hyperparameters used for each algorithm, which largely follow prior work. For consistency and computational reasons, we relied on default hyper-parameters, despite the known sensitivity of deep RL agents to these choices \citep{ceron2024on}.

\begin{table}[h]
\centering
\caption{PQN Hyperparameters}
\label{tab:pqn_hyperparameters}
\begin{tabular}{ll}
\toprule
\textbf{Hyperparameter} & \textbf{Value / Description} \\
\midrule
Learning rate & 2.5e-4 \\
Anneal lr & False (no learning rate annealing) \\
Num envs & 128 (parallel environments) \\
Num steps & 32 (steps per rollout per environment) \\
Gamma & 0.99 (discount factor) \\
Num minibatches & 32 \\
Update epochs & 2 (policy update epochs) \\
Max grad norm & 10.0 (gradient clipping) \\
Start e & 1.0 (initial exploration rate) \\
End e & 0.005 (final exploration rate) \\
Exploration fraction & 0.10 (exploration annealing fraction) \\
Q lambda & 0.65 (Q($\lambda$) parameter) \\
Use ln & True (use layer normalization) \\
Activation fn & relu (activation function) \\
\bottomrule
\end{tabular}
\end{table}

\begin{table}[h]
\centering
\caption{PPO Hyperparameters}
\label{tab:ppo_hyperparameters}
\begin{tabular}{ll}
\toprule
\textbf{Hyperparameter} & \textbf{Value / Description} \\
\midrule
Learning rate & 2.5e-4 \\
Num envs & 8 \\
Num steps & 128 (steps per rollout per environment) \\
Anneal lr & True (learning rate annealing enabled) \\
Gamma & 0.99 (discount factor) \\
Gae lambda & 0.95 (GAE parameter) \\
Num minibatches & 4 \\
Update epochs & 4 \\
Norm adv & True (normalize advantages) \\
Clip coef & 0.1 (PPO clipping coefficient) \\
Clip vloss & True (clip value loss) \\
Ent coef & 0.01 (entropy regularization coefficient) \\
Vf coef & 0.5 (value function loss coefficient) \\
Max grad norm & 0.5 (gradient clipping threshold) \\
Use ln & False (no layer normalization) \\
Activation fn & relu (activation function) \\
Shared cnn & True (shared CNN between policy and value networks) \\
\bottomrule
\end{tabular}
\end{table}

\begin{table}[h]
\centering
\caption{PPO Hyperparameters for IsaacGym}
\label{tab:ppo_isaacgym_hyperparameters}
\begin{tabular}{ll}
\toprule
\textbf{Hyperparameter} & \textbf{Value / Description} \\
\midrule
Total timesteps & 30,000,000 \\
Learning rate & 0.0026 \\
Num envs & 4096 (parallel environments) \\
Num steps & 16 (steps per rollout) \\
Anneal lr & False (disable learning rate annealing) \\
Gamma & 0.99 (discount factor) \\
Gae lambda & 0.95 (GAE lambda) \\
Num minibatches & 2 \\
Update epochs & 4 (update epochs per PPO iteration) \\
Norm adv & True (normalize advantages) \\
Clip coef & 0.2 (policy clipping coefficient) \\
Clip vloss & False (disable value function clipping) \\
Ent coef & 0.0 (entropy coefficient) \\
Vf coef & 2.0 (value function loss coefficient) \\
Max grad norm & 1.0 (max gradient norm) \\
Use ln & False (no layer normalization) \\
Activation fn & relu (activation function) \\
\bottomrule
\end{tabular}
\end{table}

\begin{table}[h]
\centering
\caption{DQN Hyperparameters}
\label{tab:dqn_hyperparameters}
\begin{tabular}{ll}
\toprule
\textbf{Hyperparameter} & \textbf{Value / Description} \\
\midrule
Learning rate & 1e-4 \\
Num envs & 1 \\
Buffer size & 1,000,000 (replay memory size) \\
Gamma & 0.99 (discount factor) \\
Tau & 1.0 (target network update rate) \\
Target network frequency & 1000 (timesteps per target update) \\
Batch size & 32 \\
Start e & 1.0 (initial exploration epsilon) \\
End e & 0.01 (final exploration epsilon) \\
Exploration fraction & 0.10 (fraction of total timesteps for decay) \\
Learning starts & 80,000 (timesteps before training starts) \\
Train frequency & 4 (training frequency) \\
Use ln & False (no layer normalization) \\
Activation fn & relu (activation function) \\
\bottomrule
\end{tabular}
\end{table}

\begin{table}[h]
\centering
\caption{Rainbow Hyperparameters}
\label{tab:rainbow_hyperparameters}
\begin{tabular}{ll}
\toprule
\textbf{Hyperparameter} & \textbf{Value / Description} \\
\midrule
Learning rate & 6.25e-5 \\
Num envs & 1 \\
Buffer size & 1,000,000 (replay memory size) \\
Gamma & 0.99 (discount factor) \\
Tau & 1.0 (target network update rate) \\
Target network frequency & 8000 (timesteps per target update) \\
Batch size & 32 \\
Start e & 1.0 (initial exploration epsilon) \\
End e & 0.01 (final exploration epsilon) \\
Exploration fraction & 0.10 (fraction of total timesteps for decay) \\
Learning starts & 80,000 (timesteps before training starts) \\
Train frequency & 4 (training frequency) \\
N step & 3 (n-step Q-learning horizon) \\
Prioritized replay alpha & 0.5 \\
Prioritized replay beta & 0.4 \\
Prioritized replay eps & 1e-6 \\
N atoms & 51 (number of atoms in distributional RL) \\
V min & -10 (value distribution lower bound) \\
V max & 10 (value distribution upper bound) \\
Use ln & False (no layer normalization) \\
Activation fn & relu (activation function) \\
\bottomrule
\end{tabular}
\end{table}

\begin{table}[h]
\centering
\caption{Image Classification Hyperparameters (CIFAR-10)}
\label{tab:cifar10_hyperparameters}
\begin{tabular}{ll}
\toprule
\textbf{Hyperparameter} & \textbf{Value} \\
\midrule
Batch size & 256 \\
Epochs & 100 \\
Learning rate & 0.00025 \\
\bottomrule
\end{tabular}
\end{table}

\begin{table}[h]
\centering
\caption{SAC Hyperparameters}
\label{tab:sac_hyperparameters}
\begin{tabular}{ll}
\toprule
\textbf{Hyperparameter} & \textbf{Value / Description} \\
\midrule
Critic block type & SimBa \\
Critic num blocks               & \{2, 4, 6, 8\} \\
Critic hidden dim               & \{512, 1024, 1536, 2048\} \\
Target critic momentum ($\tau$) & 5e-3 \\
Actor block type & SimBa \\
Actor num blocks                & \{1, 2, 3, 4\} \\
Actor hidden dim                & \{128, 256, 384, 512\}  \\
Initial temperature ($\alpha_0$) & 1e-2 \\
Temperature learning rate       & 1e-4 \\
Target entropy ($\mathcal{H}^*$) & $|\mathcal{A}|/2$ \\
Batch size                      & 256 \\
Optimizer                       & \{AdamW, Kron\}\\
AdamW's learning rate & 1e-4 \\
Kron's learning rate & 5e-5 \\
Optimizer momentum ($\beta_1$, $\beta_2$) & (0.9, 0.999) \\
Weight decay ($\lambda$)        & 1e-2 \\
Discount ($\gamma$)             & Heuristic \\
Replay ratio                    & 2 \\
Clipped Double Q                & False\\

\bottomrule
\end{tabular}
\end{table}

\begin{table}[h]
\centering
\caption{DDPG Hyperparameters}
\label{tab:ddpg_hyperparameters}
\begin{tabular}{ll}
\toprule
\textbf{Hyperparameter} & \textbf{Value / Description} \\
\midrule
Critic block type & SimBa \\
Critic num blocks               & \{2, 4, 6, 8\} \\
Critic hidden dim               & \{512, 1024, 1536, 2048\} \\
Critic learning rate            & 1e-4 \\
Target critic momentum ($\tau$) & 5e-3 \\
Actor block type & SimBa \\
Actor num blocks                & \{1, 2, 3, 4\} \\
Actor hidden dim                & \{128, 256, 384, 512\}  \\
Actor learning rate             & 1e-4 \\
Exploration noise               & $\mathcal{N}(0, {0.1}^2)$ \\
Batch size                      & 256 \\
Optimizer                       & \{AdamW, Kron\}\\
AdamW's learning rate & 1e-4 \\
Kron's learning rate & 5e-5 \\
Optimizer momentum ($\beta_1$, $\beta_2$) & (0.9, 0.999) \\
Weight decay ($\lambda$)        & 1e-2 \\
Discount ($\gamma$)             & Heuristic \\
Replay ratio                    & 2 \\
Clipped Double Q                & False\\

\bottomrule
\end{tabular}
\end{table}

\clearpage
\section{Compute Details}
\label{app:compute_details}
All experiments were conducted on a single-GPU setup using an NVIDIA RTX 8000, 12 CPU workers, and 50GB of RAM.

\begin{table}[h]
\centering
\caption{\textbf{Training times across model scales for two optimizers} K-FAC shows increased cost as depth and width grow.}
\label{tab:exp_times}
\begin{tabular}{@{}lll c@{}}
\toprule
\textbf{Depth} & \textbf{Width} & \textbf{Optimizer} & \textbf{Time} \\
\midrule
\multicolumn{4}{l}{\textit{RAdam}} \\
Small  & Small  & Adam & 51m    \\
Small  & Medium & Adam & 53m    \\
Small  & Large  & Adam & 57m    \\
Medium & Small  & Adam & 1h 4m  \\
Medium & Medium & Adam & 1h 10m \\
Medium & Large  & Adam & 1h 11m \\
Large  & Small  & Adam & 1h 18m \\
Large  & Medium & Adam & 1h 18m \\
Large  & Large  & Adam & 1h 27m \\
\midrule
\multicolumn{4}{l}{\textit{Kron}} \\
Small  & Small  & Kron & 1h 59m \\
Small  & Medium & Kron & 2h 27m \\
Small  & Large  & Kron & 3h 38m \\
Medium & Small  & Kron & 2h 44m \\
Medium & Medium & Kron & 3h 32m \\
Medium & Large  & Kron & 5h 59m \\
Large  & Small  & Kron & 3h 27m \\
Large  & Medium & Kron & 4h 36m \\
Large  & Large  & Kron & 7h 42m \\
\bottomrule
\end{tabular}
\end{table}

\subsection{Results on the Full ALE}
\label{app:fullALE}

\begin{figure}[!h]
\centering
\includegraphics[width=0.85\linewidth]{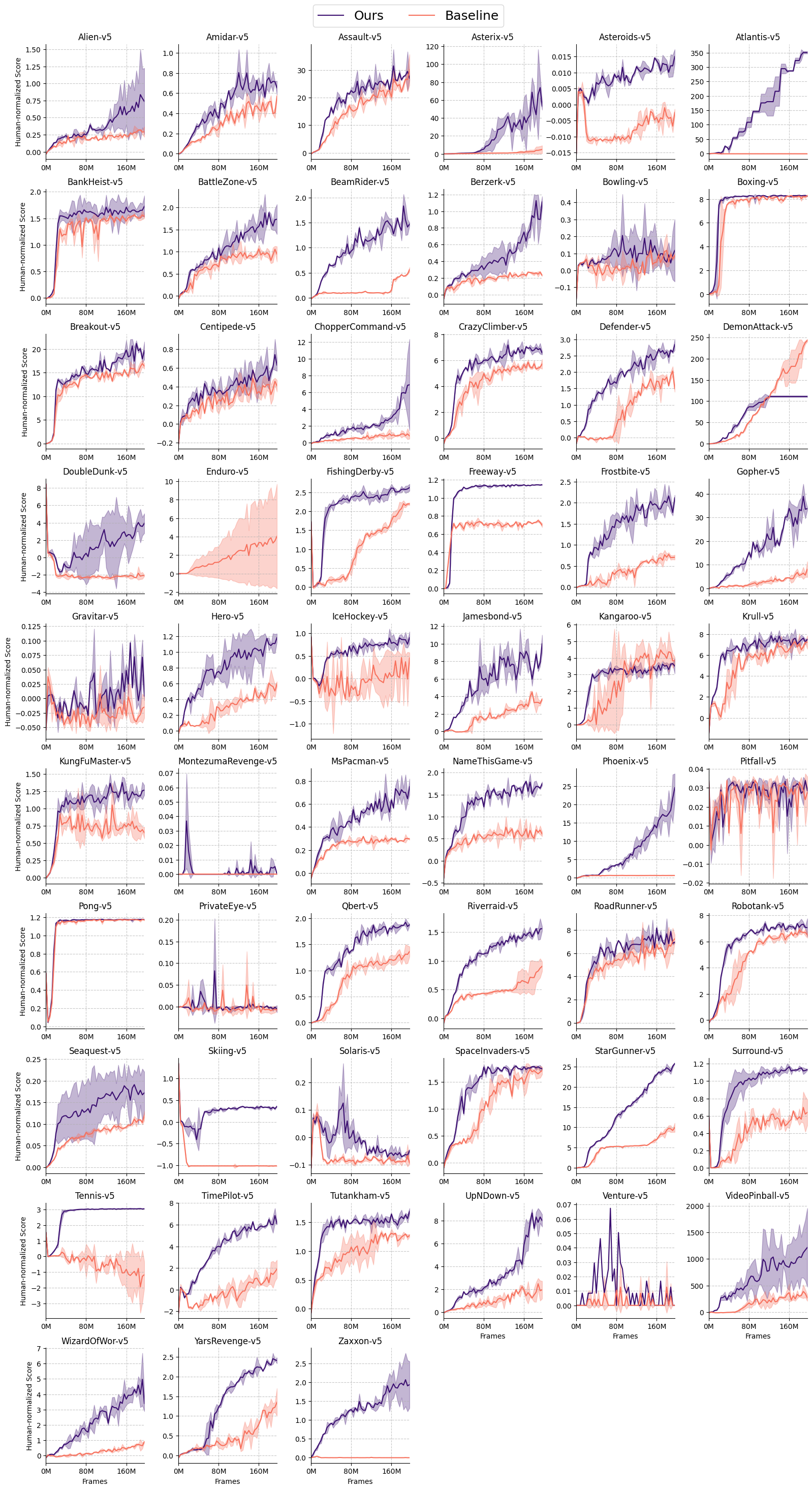}
\caption{Mean human-normalized score on the full ALE suite, comparing the baseline PQN agent (light curves) with the augmented agent using our combined gradient interventions (dark curves).}
\label{fig:pqn_fullAtari}
\end{figure}

\begin{figure}[!h]
\centering
\includegraphics[width=0.85\linewidth]{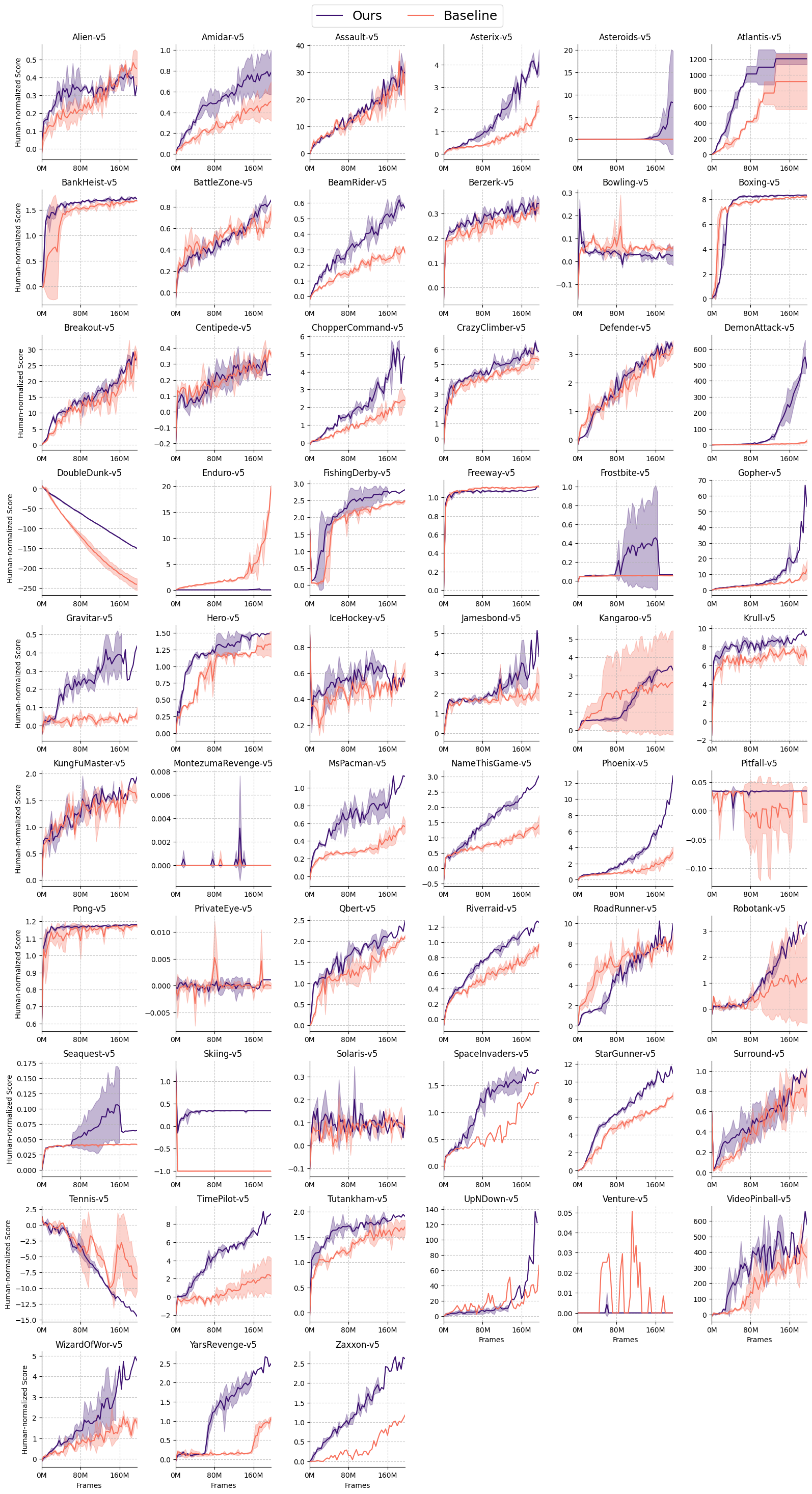}
\caption{Mean human-normalized score on the full ALE suite, comparing the baseline PPO agent (light curves) with the augmented agent using our combined gradient interventions (dark curves).}
\label{fig:ppo_fullAtari}
\end{figure}

In this section, we provide the full training curves corresponding to the aggregated results shown in \autoref{sec:puttingItAllTogether}, where we evaluate the performance of the PQN and PPO agents on the full set of environments from the ALE after applying our two proposed gradient interventions. The per-environment learning curves are presented in \autoref{fig:pqn_fullAtari} for PQN and \autoref{fig:ppo_fullAtari} for PPO. We follow the experimental protocol of \citet{obando2023small, ceron2024mixtures, agarwal2021deep}, running each experiment with three random seeds.

\clearpage

\subsection{Simba on DMC}
\label{sec:simbaFullResults}

In this section, we present the full results accompanying the experiments combining Simba \citep{lee2025simba} with our proposed gradient interventions, as introduced in \autoref{sec:puttingItAllTogether}. For these experiments, we retain Simba's original architectural choices but replace the AdamW optimizer with Kron.

We compare Simba using both SAC and DDPG as the underlying RL algorithms. While SAC generally outperforms DDPG, we consistently observe that scaling depth and width, either independently or jointly, leads to a degradation in performance with Simba. However, this degradation is mitigated, and in many cases reversed, when using the Kron optimizer, resulting in improved performance as model capacity increases.

The following figures illustrate these findings:

\begin{figure}[!h]
    \includegraphics[width=\linewidth]{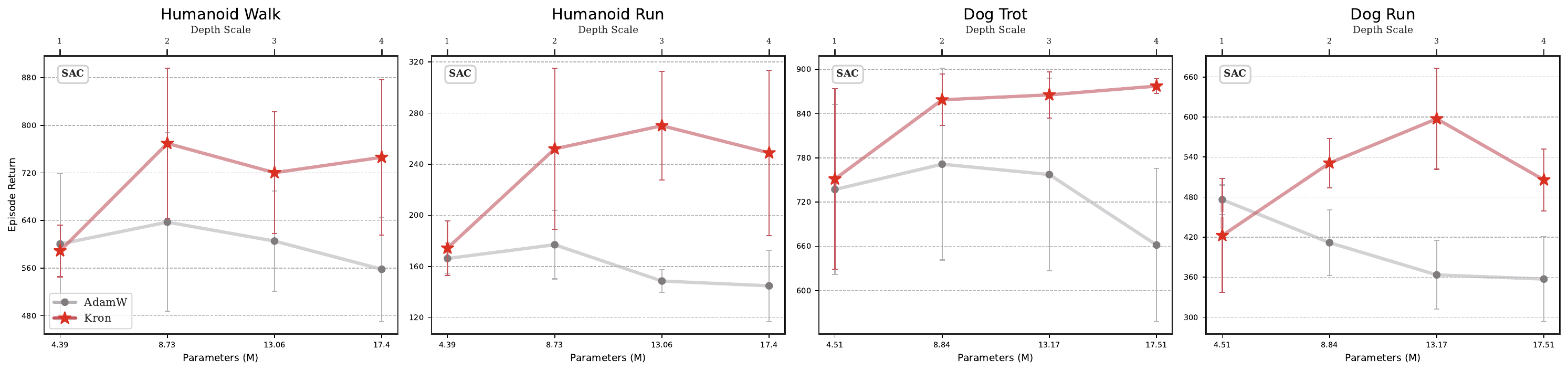}
    \caption{SAC scaling depth}
    \label{fig:simba-sac-depth}
\end{figure}

\begin{figure}[!h]
    \includegraphics[width=\linewidth]{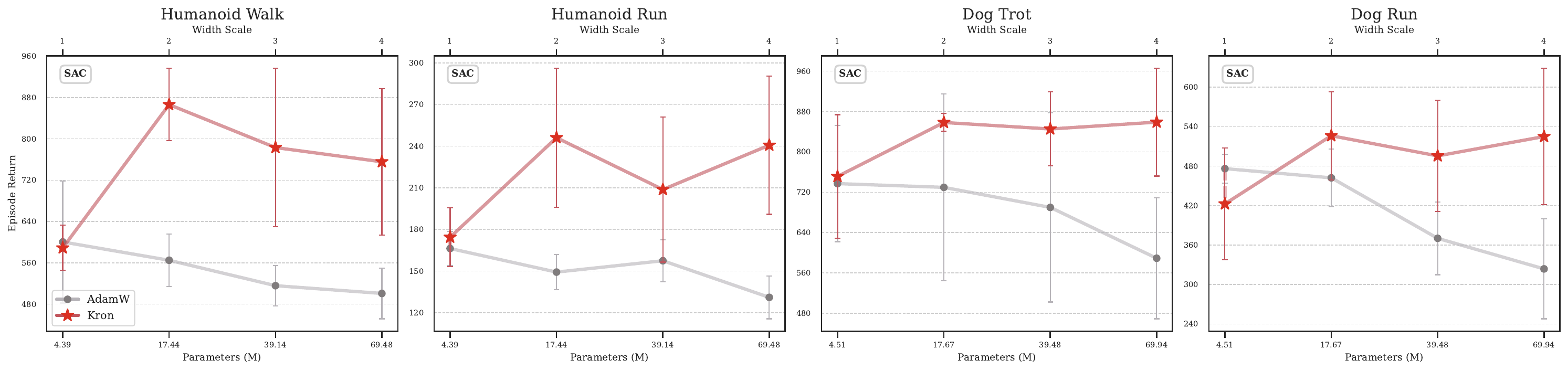}
    \caption{SAC scaling width}
    \label{fig:simba-sac-width}
\end{figure}

\begin{figure}[!h]
    \includegraphics[width=\linewidth]{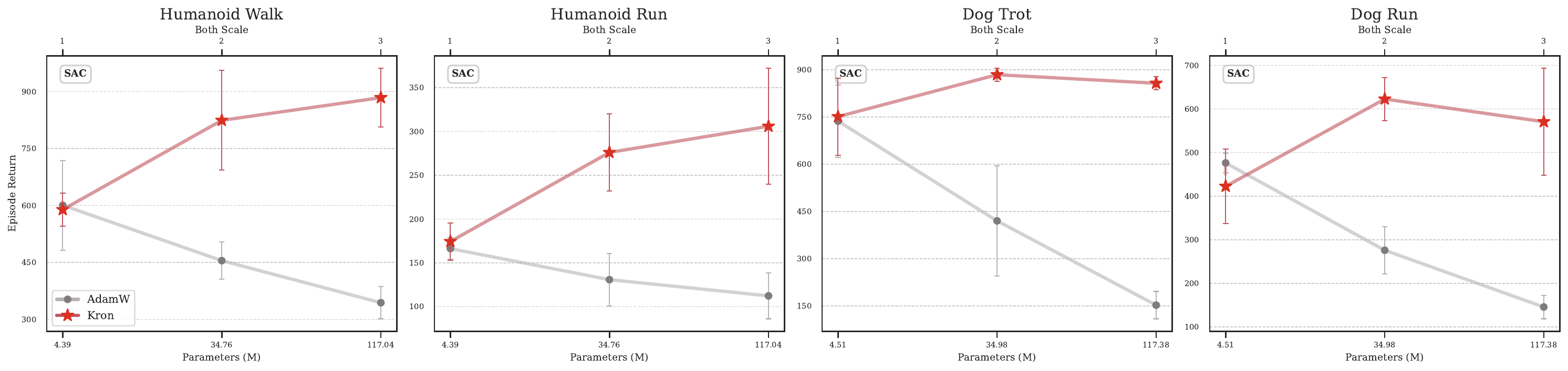}
    \caption{SAC scaling both depth and width}
    \label{fig:simba-sac-both}
\end{figure}

\begin{figure}[!h]
    \includegraphics[width=\linewidth]{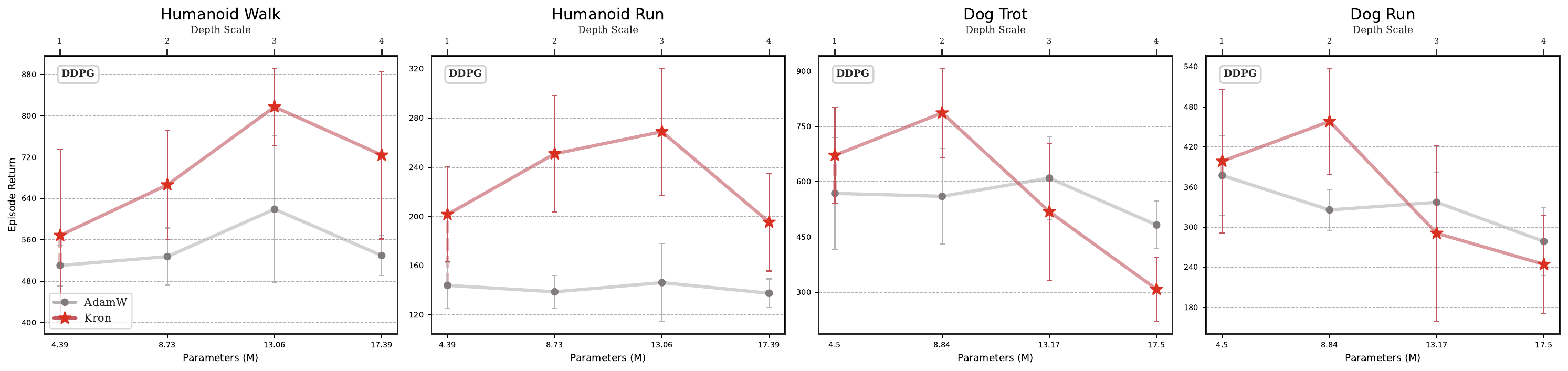}
    \caption{DDPG scaling depth}
    \label{fig:simba-ddpg-depth}
\end{figure}

\begin{figure}[!h]
    \includegraphics[width=\linewidth]{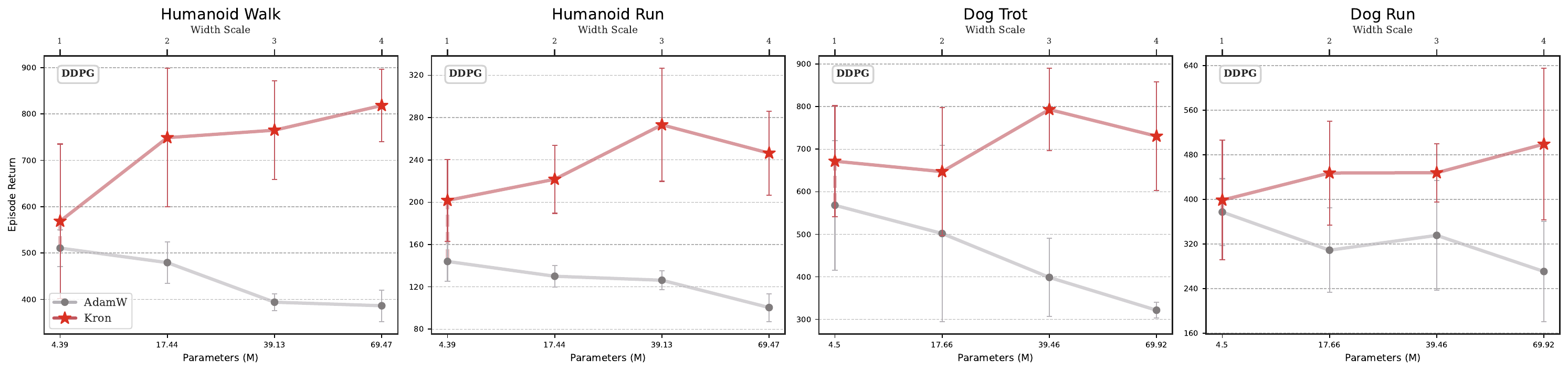}
    \caption{DDPG scaling width}
    \label{fig:simba-ddpg-width}
\end{figure}

\begin{figure}[!h]
    \includegraphics[width=\linewidth]{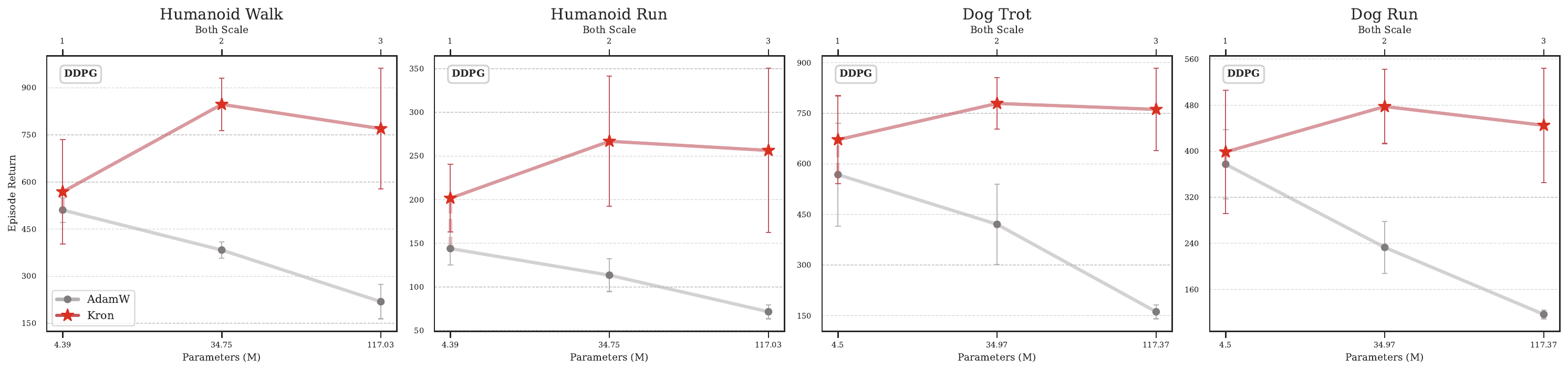}
    \caption{DDPG scaling both depth and width}
    \label{fig:simba-ddpg-both}
\end{figure}



\clearpage
\newpage
\section*{NeurIPS Paper Checklist}

\begin{enumerate}

\item {\bf Claims}
    \item[] Question: Do the main claims made in the abstract and introduction accurately reflect the paper's contributions and scope?
    \item[] Answer: \answerYes{} 
    \item[] Justification: The claims in the abstract and introduction are grounded in empirical evidence gathered under controlled settings, spanning both supervised and reinforcement learning regimes. We identify and characterize gradient pathologies that emerge with increasing model scale. The proposed gradient-based interventions are evaluated in a modular fashion throughout the paper. Furthermore, the scope of our conclusions is clearly delineated, with detailed descriptions of the specific algorithms and environments studied, ensuring alignment between claims and contributions. While our claims are carefully scoped, our extensive experimentation provides preliminary evidence suggesting that the uncovered phenomena may generalize to broader applications in deep learning.

\item {\bf Limitations}
    \item[] Question: Does the paper discuss the limitations of the work performed by the authors?
    \item[] Answer: \answerYes{} 
    \item[] Justification: See \autoref{sec:discussion}.

\item {\bf Theory assumptions and proofs}
    \item[] Question: For each theoretical result, does the paper provide the full set of assumptions and a complete (and correct) proof?
    \item[] Answer: \answerYes{} 
    \item[] Justification: The paper does not include novel theoretical results, but presents well-established claims in Section \ref{sec:preliminaries}. 

    \item {\bf Experimental result reproducibility}
    \item[] Question: Does the paper fully disclose all the information needed to reproduce the main experimental results of the paper to the extent that it affects the main claims and/or conclusions of the paper (regardless of whether the code and data are provided or not)?
    \item[] Answer: \answerYes{} 
    \item[] Justification: The experimental setup is thoroughly documented in each section presenting results. We specify the number of independent runs, the methods used for aggregating results, and provide detailed descriptions of the algorithms, hyperparameters, and tasks. This level of disclosure ensures that the main experimental findings and conclusions can be independently reproduced.

\item {\bf Open access to data and code}
    \item[] Question: Does the paper provide open access to the data and code, with sufficient instructions to faithfully reproduce the main experimental results, as described in supplemental material?
    \item[] Answer: \answerNo{} 
    \item[] Justification: Although the data and code will not accompany the submission of this work, they will be released upon publication.

\item {\bf Experimental setting/details}
    \item[] Question: Does the paper specify all the training and test details (e.g., data splits, hyperparameters, how they were chosen, type of optimizer, etc.) necessary to understand the results?
    \item[] Answer: \answerYes{} 
    \item[] Justification: We provide all essential details of the experimental setup in the main paper, including key training configurations. Additional specifics, such as hyperparameter values and environment settings, are included in the Appendix (\autoref{app:hyperparameters}, \autoref{app:environment_details}), ensuring that readers have access to all necessary information to understand and interpret the results.

\item {\bf Experiment statistical significance}
    \item[] Question: Does the paper report error bars suitably and correctly defined or other appropriate information about the statistical significance of the experiments?
    \item[] Answer: \answerYes{} 
    \item[] Justification: Throughout the paper, we report statistical variability using confidence intervals and standard deviations in training curves, error bars in bar plots, and percentiles in boxplots. We also clearly state the aggregation methods used for each plot (e.g., mean, median, inter-quantile mean), ensuring transparency and appropriate interpretation of the statistical significance of our results.

\item {\bf Experiments compute resources}
    \item[] Question: For each experiment, does the paper provide sufficient information on the computer resources (type of compute workers, memory, time of execution) needed to reproduce the experiments?
    \item[] Answer: \answerYes{} 
    \item[] Justification: This information is provided in Appendix \ref{app:compute_details}.
    
\item {\bf Code of ethics}
    \item[] Question: Does the research conducted in the paper conform, in every respect, with the NeurIPS Code of Ethics \url{https://neurips.cc/public/EthicsGuidelines}?
    \item[] Answer: \answerYes{} 
    \item[] Justification: The authors preserved anonymity during the submission of this work.

\item {\bf Broader impacts}
    \item[] Question: Does the paper discuss both potential positive societal impacts and negative societal impacts of the work performed?
    \item[] Answer: \answerYes{} 
    \item[] Justification: See Appendix \ref{app:broader_impact}.
    
\item {\bf Safeguards}
    \item[] Question: Does the paper describe safeguards that have been put in place for responsible release of data or models that have a high risk for misuse (e.g., pretrained language models, image generators, or scraped datasets)?
    \item[] Answer: \answerNA{} 
    \item[] Justification: The paper poses no such risks.

\item {\bf Licenses for existing assets}
    \item[] Question: Are the creators or original owners of assets (e.g., code, data, models), used in the paper, properly credited and are the license and terms of use explicitly mentioned and properly respected?
    \item[] Answer: \answerNA{} 
    \item[] Justification: The paper does not use existing assets.

\item {\bf New assets}
    \item[] Question: Are new assets introduced in the paper well documented and is the documentation provided alongside the assets?
    \item[] Answer: \answerNA{} 
    \item[] Justification: The paper does not release new assets.

\item {\bf Crowdsourcing and research with human subjects}
    \item[] Question: For crowdsourcing experiments and research with human subjects, does the paper include the full text of instructions given to participants and screenshots, if applicable, as well as details about compensation (if any)? 
    \item[] Answer: \answerNA{} 
    \item[] Justification: The paper does not involve crowdsourcing nor research with human subjects.

\item {\bf Institutional review board (IRB) approvals or equivalent for research with human subjects}
    \item[] Question: Does the paper describe potential risks incurred by study participants, whether such risks were disclosed to the subjects, and whether Institutional Review Board (IRB) approvals (or an equivalent approval/review based on the requirements of your country or institution) were obtained?
    \item[] Answer: \answerNA{} 
    \item[] Justification: The paper does not involve crowdsourcing nor research with human subjects.

\item {\bf Declaration of LLM usage}
    \item[] Question: Does the paper describe the usage of LLMs if it is an important, original, or non-standard component of the core methods in this research? Note that if the LLM is used only for writing, editing, or formatting purposes and does not impact the core methodology, scientific rigorousness, or originality of the research, declaration is not required.
    \item[] Answer: \answerNA{} 
    \item[] Justification: The core method development in this research does not involve LLMs as any important, original, or non-standard components.

\end{enumerate}

\end{document}